\documentclass[lettersize,journal]{IEEEtran}
\usepackage[table]{xcolor}
\usepackage{cite}   
\makeatletter
\let\NAT@parse\undefined
\makeatother
\usepackage[dvipsnames]{xcolor}
\usepackage{amsmath,amsfonts}
\usepackage{algorithmic}
\usepackage{algorithm}
\usepackage{array}
\usepackage[caption=false,font=normalsize,labelfont=sf,textfont=sf]{subfig}
\usepackage{textcomp}
\usepackage{stfloats}
\usepackage{url}
\usepackage{verbatim}
\usepackage{graphicx}
\usepackage[colorlinks=true,citecolor=RoyalBlue]{hyperref}
\usepackage{xcolor}
\usepackage{microtype} 
\usepackage{makecell}
\usepackage{booktabs}  % 引入 booktabs 宏包
\usepackage{bm}
\begin{document}

\title{PoseMoE: Mixture-of-Experts Network for Monocular 3D Human Pose Estimation}

%\author{\IEEEauthorblockN{Guangsheng Xu, Guoyi Zhang, Lejia Ye, Shuwei Gan, Xiaohu Zhang, and Xia Yang}
%\thanks{This paper was produced by the IEEE Publication Technology Group. They are in Piscataway, NJ.}% <-this % stops a space
%\thanks{Manuscript received April 19, 2021; revised August 16, 2021.}}

\author{Mengyuan Liu, Jiajie Liu, Jinyan Zhang, Wenhao Li, Junsong Yuan

\thanks{This work was supported by National Key Research and Development Program of China under Grant 2024YFB4709800, in part by Guangdong S$\&$T Program under Grant 2024B0101050002, and in part by Shenzhen Innovation in Science and Technology Foundation for The Excellent Youth Scholars under Grant RCYX20231211090248064. (Corresponding Author: Jiajie Liu)}% <-this % stops a space
\thanks{Mengyuan Liu, Jiajie Liu, and Jinyan Zhang are with the State Key Laboratory of General Artificial Intelligence, Peking University, Shenzhen Graduate School, Shenzhen, Guangdong 518055, China (e-mail: liumengyuan@pku.edu.cn; liujiajie@stu.pku.edu.cn; zhangjinyan@stu.pku.edu.cn). Wenhao Li is with the College of Computing and Data Science, Nanyang Technological University, Singapore 639798 (e-mail: wenhao.li@ntu.edu.sg). Junsong Yuan is with the Department of Computer Science and Engineering, University at Buffalo, State University of New York, Buffalo, NY 14228 USA (e-mail: jsyuan@buffalo.edu).}}

% \markboth{Submission to IEEE Transactions on Image Processing}{}%
\markboth{IEEE Transactions on Image Processing}{}%



\maketitle

\begin{abstract}
% %
% The lifting-based methods have dominated monocular 3D human pose estimation by leveraging well-detected 2D poses as intermediate representations. 
% %
% However, it neglects different initial states between 2D pose and depth. 
% % 
% The initial state of the 2D pose is well-detected, but the depth is unknown and needs to be learned from scratch.
% %
% The lifting-based methods encode the well-detected 2D pose and unknown depth in an entangled feature space, explicitly introducing depth uncertainty to the well-detected 2D pose.
% %
% To address this limitation, we present a Mixture-of-Experts network for monocular 3D pose estimation named \textbf{PoseMoE}. 
% %
% Our approach introduces: (1) A mixture-of-expert network where specialized expert modules refine the well-detected 2D pose features and learn the depth features. This mixture-of-expert design reduces the explicit influence of uncertain depth features on 2D pose features.
% %
% (2) A cross-expert knowledge aggregation module is proposed to aggregate cross-expert spatio-temporal contextual information. This step enhances features through bidirectional mapping between 2D pose and depth.
% %
% Extensive experiments show that our proposed PoseMoE outperforms the conventional lifting-based methods with fewer parameters on two widely used datasets: Human3.6M and MPI-INF-3DHP. 

The lifting-based methods have dominated monocular 3D human pose estimation by leveraging detected 2D poses as intermediate representations. 
The 2D component of the final 3D human pose benefits from the detected 2D poses, whereas its depth counterpart must be estimated from scratch. 
The lifting-based methods encode the detected 2D pose and unknown depth in an entangled feature space, explicitly introducing depth uncertainty to the detected 2D pose, thereby limiting overall estimation accuracy. 
This work reveals that the depth representation is pivotal for the estimation process. 
Specifically, when depth is in an initial, completely unknown state, jointly encoding depth features with 2D pose features is detrimental to the estimation process.
In contrast, when depth is initially refined to a more dependable state via network-based estimation, encoding it together with 2D pose information is beneficial.
To address this limitation, we present a Mixture-of-Experts network for monocular 3D pose estimation named \textbf{PoseMoE}. 
Our approach introduces: (1) A mixture-of-experts network where specialized expert modules refine the well-detected 2D pose features and learn the depth features. This mixture-of-experts design disentangles the feature encoding process for 2D pose and depth, therefore reducing the explicit influence of uncertain depth features on 2D pose features.
(2) A cross-expert knowledge aggregation module is proposed to aggregate cross-expert spatio-temporal contextual information. This step enhances features through bidirectional mapping between 2D pose and depth.
Extensive experiments show that our proposed PoseMoE outperforms the conventional lifting-based methods on three widely used datasets: Human3.6M, MPI-INF-3DHP, and 3DPW.

% Code is available at \url{https://anonymous.4open.science/r/PoseMoE}.
\end{abstract}


\section{Introduction}
\label{sec:intro}

\IEEEPARstart{M}{onocular} 3D human pose estimation has been a crucial problem in computer vision, which aims to locate the 3D joint positions of a human body~\cite{ moon2020i2l, pavlakos2018ordinal, chen2021anatomy}. 
Nowadays, monocular 3D human pose estimation finds widespread applications in various scenarios, including motion prediction~\cite{ liu2021motion, liu2022investigating, wang2023dynamic,wang2024gcnext,cui2024hybrid}, action recognition~\cite{zhang2022unsupervised,wang2024merge,ren2024survey}, and human-robot interaction~\cite{ gong2022meta, ye2021collaborative}. 
Existing monocular 3D human pose estimation methods can be categorized as the end-to-end manner and lifting-based manner. 
The end-to-end approaches~\cite{kanazawa2018end, pavlakos2017coarse, sun2018integral} directly estimate the 3D pose from the input image without the intermediate 2D pose representation.
Different from the end-to-end manner, lifting-based methods~\cite{simplebaseline, liu2020comprehensive} first obtain 2D pose using 2D pose detector~\cite{newell2016stacked,chen2018cascaded} and then lift the 2D pose in image coordinate to the 3D pose in camera coordinate. 
These lifting-based methods usually outperform the end-to-end manner and dominate the monocular 3D human pose estimation.
% have been the dominant paradigm in monocular 3D human pose estimation.
Their primary advantage lies in leveraging well-detected 2D poses as a powerful intermediate representation.
Furthermore, the lightweight nature of 2D pose data enables models to effectively utilize long-term temporal clues to achieve advanced accuracy.

\begin{figure}[t] 
\centering
\includegraphics[width=1\linewidth]{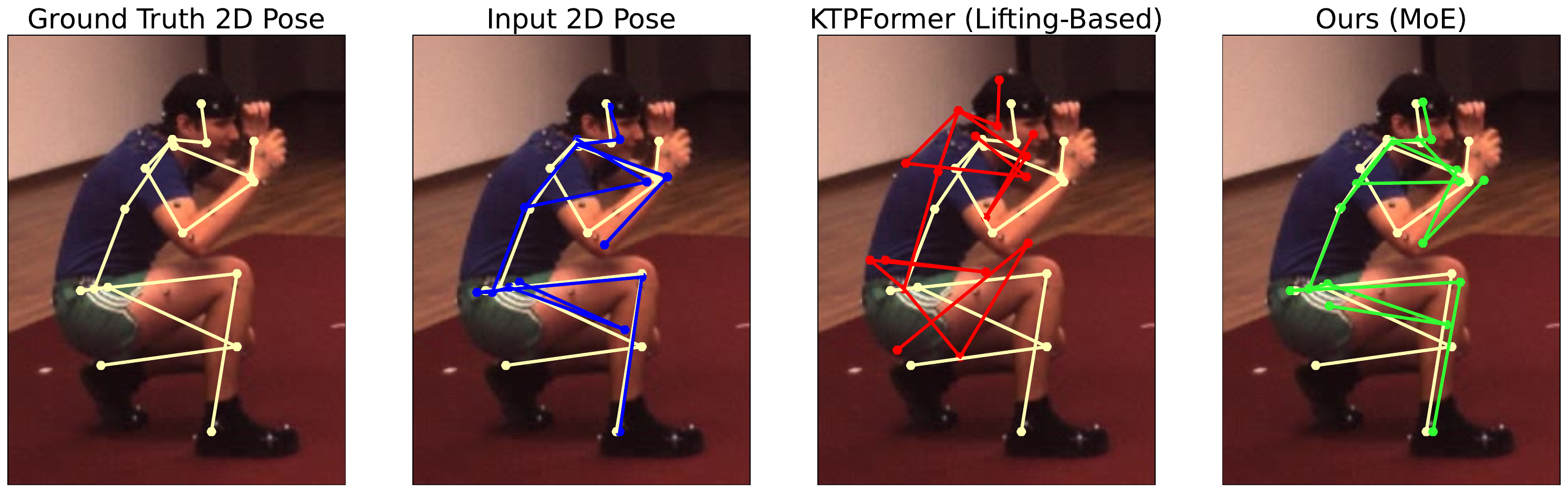}
\caption{An illustration of our motivation. We project the 2D pose in the camera coordinate (part of the output 3D pose) back to the image coordinate for comparison. The powerful lifting-based method KTPFormer~\cite{peng2024ktpformer} obtains a 2D pose worse than the input, which contradicts our intuition. In contrast, our framework obtains a 2D pose better than the input.
% Please refer to Appendix~\ref{appendix:vis} for more qualitative \textcolor{magenta}{and quantitative} comparisons.
}
\label{fig:2}
\end{figure}

\begin{figure*}[t] 
\centering
\includegraphics[width=1\linewidth]{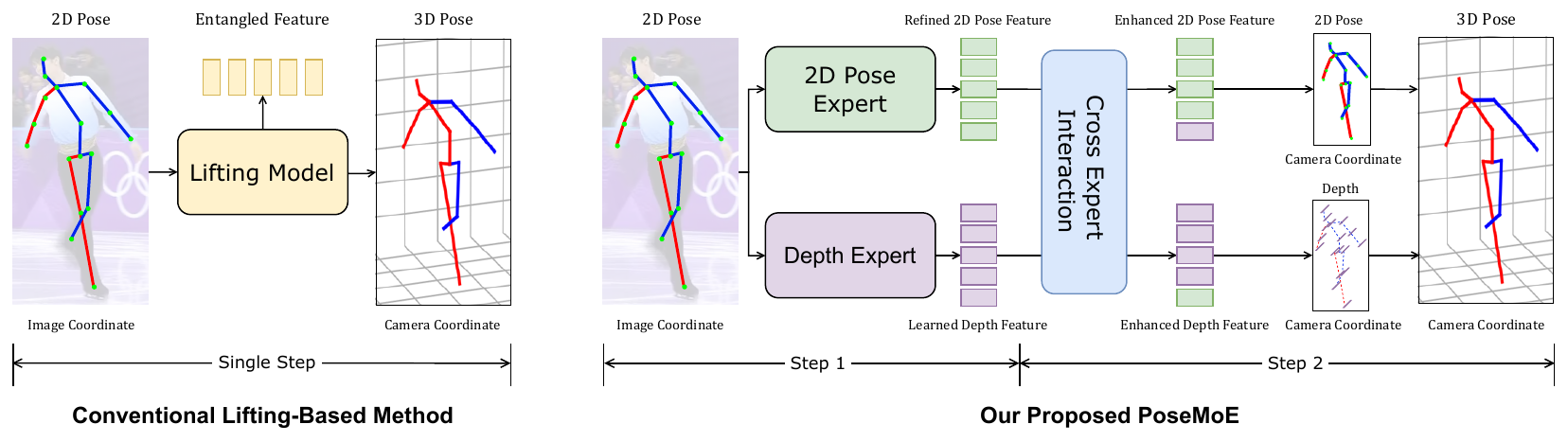}
\caption{Given a 2D pose in the image coordinate, we aim to estimate the 3D pose in the camera coordinate. Left: Conventional lifting-based methods directly project the 2D pose in an entangled feature space and regression the 3D pose from it. Right: Our proposed Mixture-of-Experts network. The 2D pose and depth features are learned separately through expert model. Then, we perform feature interaction to supplement the complementary information between 2D pose feature and depth feature. Finally, we regress the 2D pose and depth and concatenate them to obtain the final 3D pose. 
}
\label{fig:1}
\end{figure*}

\begin{figure*}[t] 
\centering
\includegraphics[width=1\linewidth]{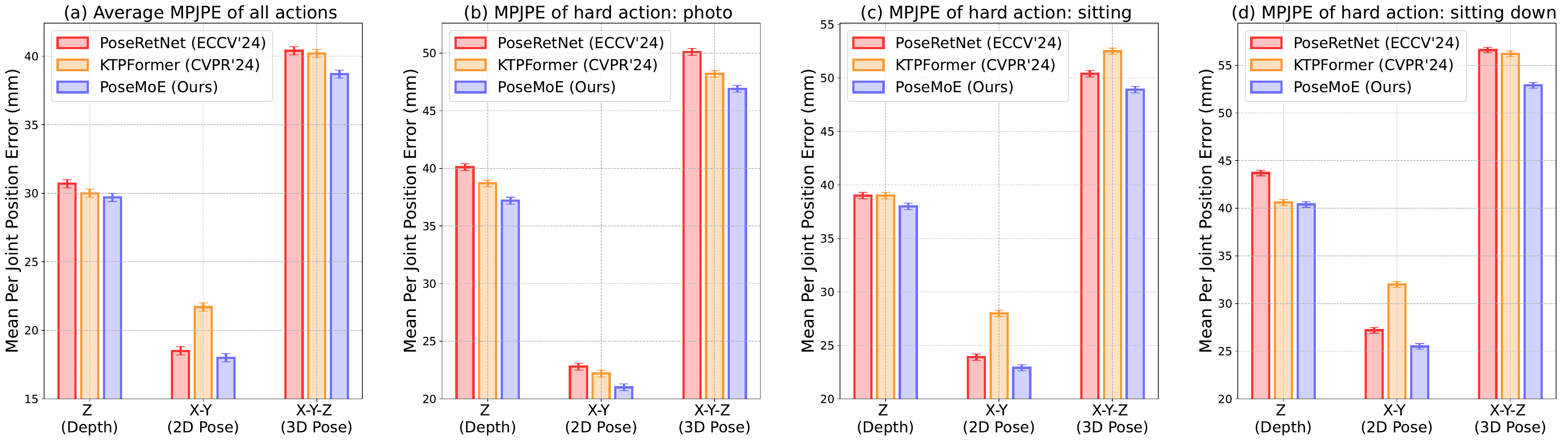}
\caption{Quantitative Comparison of Mean Per Joint Position Error (MPJPE) of different axes for all actions and three hard actions~\cite{zeng2021learning} with SOTA lifting-based methods~\cite{zheng20243d,peng2024ktpformer}. The MPJPE of the Z-axis (depth) is significantly higher than the X-Y axes (2D pose) and accounts for the majority of the overall (3D pose) MPJPE. Our proposed method achieves better results across different axes than the lifting-based framework.
}
\label{fig:3}
\end{figure*}

Recent lifting-based methods~\cite{poseformer,mhformer,mixste,yu2023gla,motionbert,peng2024ktpformer} for monocular 3D human pose estimation focus on designing various spatio-temporal encoders. 
% While these methods have achieved impressive results, their approach is still within the lifting-based framework. 
As shown in Figure~\ref{fig:1} left, they project the 2D pose into an entangled feature space and regress the 3D pose from it. 
The essence of this approach is that the network must implicitly solve two core sub-problems within a single framework: first, optimizing the input 2D pose (including denoising and coordinate transformation), and second, estimating the unknown depth from scratch.  
These conventional models overlook the fact that the 2D structural features (X, Y) are relatively reliable, while the depth features (Z) are inherently ambiguous due to monocular information loss; when fused in a single encoder, the uncertainty of Z forces the feature space to accommodate its ambiguities, thereby corrupting the accurate spatial constraints of X and Y. 
%
% It encodes the well-detected 2D pose and unknown depth features in an entangled feature space, which introduces a main limitation. 
%
% \textbf{First, the uncertainty of the depth feature may erode the 2D pose.} 
%
\textbf{It encodes the well-detected 2D pose and unknown depth in an entangled feature space, which introduces a main limitation: the high uncertainty of the depth may erode the 2D pose.}
It is well-known that the monocular 3D human pose estimation task is an ill-posed problem and inherently suffers from depth ambiguity~\cite{mhformer,ma2021context,wehrbein2021probabilistic}. 
One 2D pose possibly corresponds to multiple 3D poses, where the lifting process is inherently ambiguous~\cite{yu2021towards}. 
To validate the impact of depth uncertainty on the 2D pose, we project the 2D pose in the camera coordinate (part of the output 3D pose) back to the image coordinate and compare it with the ground truth 2D pose and input 2D pose. 
As shown in Figure~\ref{fig:2}, despite learning through multiple spatio-temporal encoders, the 2D pose of the lifting method KTPFormer~\cite{peng2024ktpformer} could be even worse than the original input 2D pose. 
This observation provides empirical evidence that conventional lifting-based methods will introduce explicit uncertainty to the 2D pose and cause erosion. 
%
% \textbf{Second, different initial states of 2D pose and depth lead to optimization differences.}
%
% For the 2D pose, we need to learn implicit depth information to construct a projection matrix to project the well-detected 2D pose from the image coordinate to the camera coordinate.
% In contrast, for the unknown depth, we need to learn it from scratch explicitly. 
%
% To provide more empirical support for the high uncertainty of depth, we conduct quantitative comparisons of Mean Per Joint Position Error (MPJPE) across different axes for different hard actions~\cite{zeng2021learning} with PoseRetNet~\cite{zheng20243d} and KTPFormer~\cite{peng2024ktpformer}. 
%
% As shown in Figure~\ref{fig:3}, the MPJPE of depth is higher than the MPJPE of the 2D pose. 
% As shown in Figure~\ref{fig:3}, the MPJPE of depth is significantly higher than the MPJPE of 2D pose and accounts for the majority of the overall MPJPE.
To provide more empirical support for the high uncertainty of depth, we conduct quantitative comparisons of Mean Per Joint Position Error across different axes for different hard actions~\cite{zeng2021learning} with PoseRetNet~\cite{zheng20243d} and KTPFormer~\cite{peng2024ktpformer} in Figure~\ref{fig:3}. This decomposition meticulously isolates the error contributions into the Z component (Depth) and the X-Y component (2D Pose). It reveals that in traditional joint-encoding methods, particularly during challenging and hard actions such as ”photo” and ”sitting”, a high Z Error (Depth) is consistently and concurrently accompanied by a significant X-Y Error (2D Pose). This robust empirical correlation provides compelling proof, unequivocally confirming that the Z-axis instability, which arises from inherent depth ambiguity, directly propagates into and compromises the X-Y structural domain, thereby reducing the accuracy of the planar features themselves.
%
% These results show optimization differences between the well-detected 2D pose and unknown depth.
These quantitative findings highlight the high uncertainty of depth compared to the well-detected 2D pose.
This raises a fundamental question: Is it necessary to solve 2D pose optimization and depth estimation synchronously and implicitly within a single network? 
Motivated by these qualitative and quantitative observations, we propose a Mixture-of-Experts network named \textbf{PoseMoE} to address this limitation. 
As shown in Figure~\ref{fig:1} right, the 2D pose and depth features are learned respectively through the expert model. 
%
% The dual-branch design brings two benefits. First, learning the features of the 2D pose and depth by different expert models avoids the explicit impact of uncertain depth features on the 2D pose. 
% %
% Second, the model parameters are not shared across different expert models, which makes the training more targeted.
% %
% Furthermore, while 2D pose and depth are intuitively and physically strongly correlated, the question remains as to why previous joint optimization approaches perform sub-optimally.
% %
% We conclude that the initial quality of depth critically affects the outcome of joint optimization. 
% %
% Specifically, when depth is in an initial, completely unknown state, jointly encoding its features with 2D pose features is detrimental. 
% %
% In contrast, when depth is first refined to a more dependable state via network estimation, encoding it together with 2D pose information is beneficial. 
% %
% Based on this insight, we introduce our second innovation: a cross-expert knowledge aggregation module designed to perform feature interaction after the two sub-problems have been initially refined by the MoE network.
% %
% This step enhances features through a bidirectional mapping between the 2D pose and depth representations.
Our method differs fundamentally from standard multi-task dual-branch networks, primarily in terms of our two contributions: PoseMoE Encoder and PoseMoE Decoder. The PoseMoE Encoder addresses the key limitation of traditional dual-branch networks, which use the same homogenous encoders and overlook the vast difference in reliability between 2D pose and depth estimation. Specifically, 3D estimation requires learning a coordinate transformation for the relatively reliable 2D pose (X, Y component) but must learn the depth (Z component) from scratch, which is inherently challenging and ambiguous. The core novelty of PoseMoE lies in its Mixture-of-Experts (MoE) Encoder architecture, which uses specialized expert design to address this discrepancy and enforce structural decoupling. The 2D Pose Expert is specifically designed to refine reliable 2D pose features, focusing only on structural constraints and preserving the robust X-Y information, thereby preventing contamination from depth uncertainty. Conversely, the Depth Expert is dedicated to learning the uncertain depth information, strategically compensating for the input's limited information entropy by leveraging learnable tokens initialized with a Gaussian distribution to acquire robust prior knowledge, which promotes stable convergence and enables the expert to derive stable depth feature representations in an isolated environment. The PoseMoE Decoder resolves the second limitation of traditional methods by implementing a strategic, conditional aggregation phase. Our analysis showed that attempting to jointly encode depth features in an initial, completely unknown state is detrimental to 2D pose quality, but aggregating them becomes highly beneficial once the depth features have been rigorously refined to a more dependable state. The Decoder implements this insight: it performs strategic, delayed knowledge aggregation via a bidirectional cross-attention mechanism only after the PoseMoE Encoder has achieved structural decoupling and refinement of the 2D pose and depth features. This specialized module establishes functional connections between the features, utilizing a powerful bidirectional mapping mechanism to dynamically enhance them. This process is critical because it enables the depth information to rigorously leverage the 2D pose's spatial dependency and temporal consistency (reducing depth's scale ambiguity), while simultaneously allowing the robust 2D poses to utilize the refined depth context to strategically correct projection errors in challenging scenarios.
As shown in Figure~\ref{fig:2}, our framework could reduce the erosion of the 2D pose caused by depth uncertainty. 
The quantitative results in Figure~\ref{fig:3} also demonstrate our methods performs favorably across different parts (2D pose and depth) of the 3D pose. 
Extensive experiments on two widely used monocular 3D human pose estimation benchmarks (i.e., Human3.6M~\cite{h36m} , MPI-INF-3DHP~\cite{3dhp} and 3DPW~\cite{von2018recovering}) demonstrate that the proposed PoseMoE outperforms conventional lifting-based methods in terms of accuracy and robustness with fewer parameters. 
The key contributions of this paper are as follows:  

\begin{itemize}
    \item We tackle an overlooked different quality between the well-detected 2D pose and the unknown depth of the lifting-based methods and present a novel mixture-of-experts network named PoseMoE to address it. 
    \item We propose a PoseMoE Encoder, which consists of a 2D pose expert and a depth expert to learn 2D pose and depth features respectively. We also propose a PoseMoE Decoder to indirectly supplement the complementary information between 2D pose and depth after expert learning.
    \item Our method achieves state-of-the-art results on Human3.6M, MPI-INF-3DHP and 3DPW datasets. These results demonstrate the potential of the mixture-of-experts for future monocular 3D human pose estimation research.
\end{itemize}

\section{Related Work}
\subsection{Monocular 3D Human Pose Estimation}
Existing methods for monocular 3D human pose estimation can be categorized as end-to-end and lifting-based. 
End-to-end approaches~\cite {kanazawa2018end, pavlakos2017coarse, sun2018integral} directly estimate the 3D pose from the input image without the intermediate 2D pose representation.
With the reliable achievement of 2D human pose detectors~\cite{chen2018cascaded, newell2016stacked, sun2019deep}, lifting-based methods~\cite{fang2018learning, simplebaseline, zhao2019semantic, liu2020comprehensive} first obtain 2D pose representations in the image and then lift the 2D joint coordinates to 3D space. 
Recently, Transformers~\cite{transformer} have been applied to various visual tasks~\cite{ViT, carion2020end}. 
For the monocular 3D human pose estimation task, PoseFormer~\cite{ poseformer} introduces transformer architecture to leverage spatial and temporal dependency. 
StridedFormer~\cite{stride} proposed a transformer module with strided temporal convolution. 
MHFormer~\cite{mhformer} addresses the depth ambiguity by learning multiple pose hypotheses and 
MixSTE~\cite{mixste} constructs a mixed spatiotemporal transformer to capture the temporal motion of different body joints.
P-STMO~\cite{pstmo} is the first approach introducing the pre-training technique to 3D human pose estimation.
PoseFormerV2~\cite{poseformerv2} improves PoseFormer by utilizing a frequency-domain representation of input joint sequences.
STCFormer~\cite{tang20233d} decomposed spatio-temporal attention and integrated the structure-enhanced positional embedding.
CA-PF~\cite{zhao2023contextaware} presents a network that leverages readily available visual representations.
On the other hand, MotionBERT~\cite{motionbert} trains a unified model for multiple downstream tasks.
In~\cite{peng2024ktpformer}, KTPFormer uses two prior attention modules to facilitate pose estimation.
Moreover, MotionAGFormer~\cite{motionagformer2024} using two parallel transformer and GCNFormer streams to better learn the underlying 3D structure.
%
% However, these methods are developed within the conventional lifting-based framework.
%, overlooking the different initial states between well-detected 2D pose and unknown depth. 
% In contrast, we propose a mixture-of-experts to estimate 3D human pose. 
There are also large-scale models tailored to 3D vision or human vision research domains, such as the 3D vision foundation model 3D-LFM~\cite{dabhi20243d} and the human vision foundation model Sapiens~\cite{khirodkar2024sapiens}. These large-scale models are capable of providing more robust and generalizable prior features to facilitate the 3D human pose estimation.
However, previous transformer-based lifting methods focus on spatiotemporal modeling and typically process all features within a single Transformer encoder. They do not explicitly address the entanglement between 2D pose features and depth features. In contrast, the core advantage of PoseMoE lies in its Mixture of Experts architecture, which assigns 2D pose features and depth features to separate, specialized expert modules for disentangled encoding.

\subsection{Mixture-of-Experts Models.} The Mixture-of-Experts (MoE) method has been proposed to increase the capacity of a deep neural network without raising computational costs. The MoE method activates only a subset of parameters for each input, with these active parameters referred to as experts. \cite{shazeer2017outrageously} introduces an MoE layer between LSTM layers, achieving impressive results in language modeling and machine translation benchmarks. Subsequently, the MoE layer is incorporated into the transformer architecture as a replacement for the feed-forward network layers. Switch Transformer~\cite{fedus2022switch} simplifies the gating by selecting only the Top-1 expert per token, achieving better scaling compared to previous methods. Gshard~\cite{lepikhin2020gshard} improves the Top-2 expert routing strategy and significantly improves machine translation across 100 languages. Besides, BASE layer~\cite{lewis2021base}, HASH layer~\cite{roller2021hash}, and Expert Choice~\cite{zhou2022mixture} explore ways to optimize MoE models for full utilization of their capacity. Recently, DeepseekMoE~\cite{dai2024deepseekmoe} and XMoE~\cite{yang2024enhancing} introduce fine-grained expert segmentation in MoE architectures. PEER~\cite{he2024mixture} expands the number of experts to one million, while LLaMA-MoE~\cite{zhu2024llama} proposes pruning the MoE model from a dense model.
We are the first that introduce the MoE architecture, which can be creatively leveraged to resolve the feature entanglement problem between 2D pose and depth, to 3D human pose estimation

\section{Rethinking Lifting-based Monocular 3D Human Pose Estimation}

Since SimpleBaseline~\cite{simplebaseline} proposes the 2D-to-3D lifting framework, numerous methods~\cite{videopose,mixste,zheng2020joint,pstmo,xue2022boosting,zhang2021learning,motionbert,wu2021limb,peng2024ktpformer,hassan2023regular,mhformer,poseformerv2,liu2025tcpformer,wang2024skeleton} have been developed within this framework.
These lifting-based methods usually outperform the end-to-end manner~\cite {kanazawa2018end, pavlakos2017coarse, sun2018integral} and have been the dominant paradigm in monocular 3D human pose estimation for a long time. 

%What makes the lifting-based framework better than the end-to-end manner? 
The ensuing question is why lifting-based methods perform better than end-to-end approaches. 
%
% This work shows that this is mainly attributed to leveraging the 2D pose as an intermediate representation. 
% We argue that this is mainly attributed to leveraging the 2D pose as an intermediate representation. 
We argue that the success of lifting-based methods is mainly attributed to two factors.
%
% First, there exists a high relevance between 2D pose and 3D pose. 
First, leveraging the 2D pose as an intermediate representation provides a dependable foundation for the 3D estimation task.
Regressing 3D pose directly from raw images is a highly nonlinear and challenging problem~\cite{pavlakos2017coarse}. 
This difficulty also exists in 2D human pose estimation~\cite{pfister2015flowing, tompson2014joint}.
In contrast, with the widespread usage of 2D human pose detectors~\cite{chen2018cascaded, he2017mask, newell2016stacked, sun2019deep}, lifting-based methods could leverage the well-detected 2D pose, which contributes to its 3D counterpart and make network training easy.
Second, the 2D pose is exceptionally lightweight compared to raw image data, which allows for the effective utilization of long-term temporal clues to address occlusion and enhance accuracy.(e.g., 243 frames for MixSTE~\cite{mixste} and KTPFormer~\cite{peng2024ktpformer}; large as 351 frames for MHFormer~\cite{mhformer})

Once we have a well-detected 2D pose, lifting it directly to 3D space is natural and simple. 
However, these lifting-based methods neglect different initial states between 2D pose and depth and encode the well-detected 2D pose features and unknown depth features in an entangled feature space.
This leads to the fact that despite these methods~\cite{motionbert,peng2024ktpformer,mhformer} striving to design various encoders to leverage the well-detected 2D pose, the 2D pose itself is inevitably eroded by the uncertainty of depth features (see Figure~\ref{fig:2}). 
%
% Once we have a well-detected 2D pose, lifting it directly to 3D space is natural and simple. 
%
%However, is the lifting-based framework the perfect way to obtain the 3D pose? 
%
This paper presents a mixture-of-experts that addresses the different initial states between 2D pose and depth and provides a new choice for future monocular 3D human pose estimation.

\section{Method}

\label{method}

\begin{figure*}[t] 
\centering
\includegraphics[width=1\linewidth]{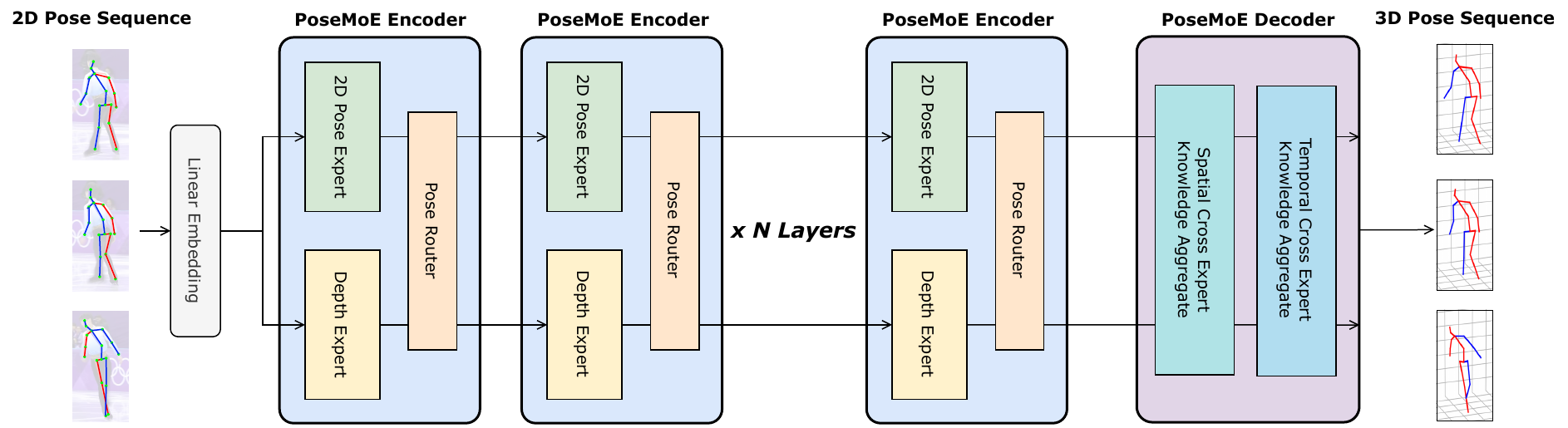}
\caption{Overview of our proposed PoseMoE network. 
We first project the 2D pose sequence to a high-dimensional feature through a linear embedding. 
These features are then handed to the PoseMoE Encoder (PME) to generate the refined 2D pose and learned depth features. 
Subsequently, we fed them into the PoseMoE Decoder (PMD) to establish the connection between 2D pose and depth, obtaining the enhanced 2D pose and depth features. 
Finally, we use two regression heads to regress the 2D pose and depth sequences, respectively, and concatenate them to obtain the 3D pose sequence.
}
\label{fig:4}
\end{figure*}

As shown in Figure~\ref{fig:4}, we first project the 2D pose sequence to a high-dimensional feature through a linear embedding. These basic features are then handed to the PoseMoE Encoder (PME) and repeated for $N$ layers.
% 2D pose expert and depth expert two task branches to generate the refined 2D pose and learned depth features. 
%
Subsequently, we fed them into the PoseMoE Decoder (PMD) to establish the connection between 2D pose and depth, obtaining the enhanced 2D pose and depth features. 
Finally, we use two regression heads to regress the 2D pose sequence and depth sequence, respectively, and concatenate them to obtain the 3D pose sequence.
In what follows, we will introduce the complete architecture of our method in detail.

\subsection{Problem Formulation}%
% \label{sub:overview}
Given a 2D pose sequence $X \in \mathbb{R}^{T \times J \times 2}$, the goal of monocular 3D human pose estimation is to estimate the 3D pose sequence $\overline{Y} \in \mathbb{R}^{T \times J \times 3}$. 
Here, $T$ refers to the number of frames, and $J$ refers to the number of joints. $ 2$ and $3$ denote the dimension of the input 2D pose and output 3D pose.

\noindent\textbf{Conventional Scheme.} Previous lifting-based methods~\cite{poseformer,poseformerv2,mixste,pstmo,motionbert,chen2023hdformer} for monocular 3D human pose estimation tend to directly projects the 2D pose in an entangled feature space and regression the 3D pose from it.
Mathematically, the formulation they followed can be summarized as:
\begin{equation}
\overline{Y} = \mathcal{F}_{Lift}(X)
\end{equation}
Where $\mathcal{F}_{Lift}$(·) denotes the 2D-to-3D lifting network.

\noindent\textbf{Proposed Scheme.}
In this paper, we introduce a novel framework that decomposes the 3D human pose estimation into 2D pose refinement and depth estimation. Specifically, we first use the multi-task learning framework $\mathcal{F}_{MTL}$(·) to obtainthe refined 2D pose $ X_{R} \in \mathbb{R}^{T \times J \times 2}$ and depth  $D \in \mathbb{R}^{T \times J \times 1}$ respectively. Then, we concatenate them to generate the  3D pose sequence $\overline{Y}$ as:
\begin{equation}
\begin{split}
\overline{Y} = Y_{2D} \oplus Y_{D}  \quad \text{where  }   Y_{2D}, Y_{D} = \mathcal{F}_{MoE}(X)
\end{split}
\end{equation}
Here $\mathcal{F}_{MoE}$(·) denotes our proposed PoseMoE network.

\subsection{PoseMoE Encoder}
The PoseMoE Encoder consists of a 2D pose expert, a depth expert, and a pose router. The 2D pose expert is responsible for refining 2D pose features, while the depth expert learns depth features from scratch. The pose router performs a coarse fusion based on the correlation between the two feature types.
We first use a linear embedding layer to project the 2D pose sequence into a high dimension to extract the general feature $F \in \mathbb{R}^{T \times J \times C}$. $C$ denotes the feature dimension.
Next, we add the learnable 2D pose position embedding and depth position embedding to $F$ to obtain the 2D pose features $F_{2D} \in \mathbb{R}^{T \times J \times C}$ and the depth features $F_{D} \in \mathbb{R}^{T \times J \times C}$ respectively.  $F_{2D}$ and $F_{D}$ are respectively fed into the 2D pose expert and the depth expert for further feature learning.
\begin{figure}[t] 
\centering
\includegraphics[width=1\linewidth]{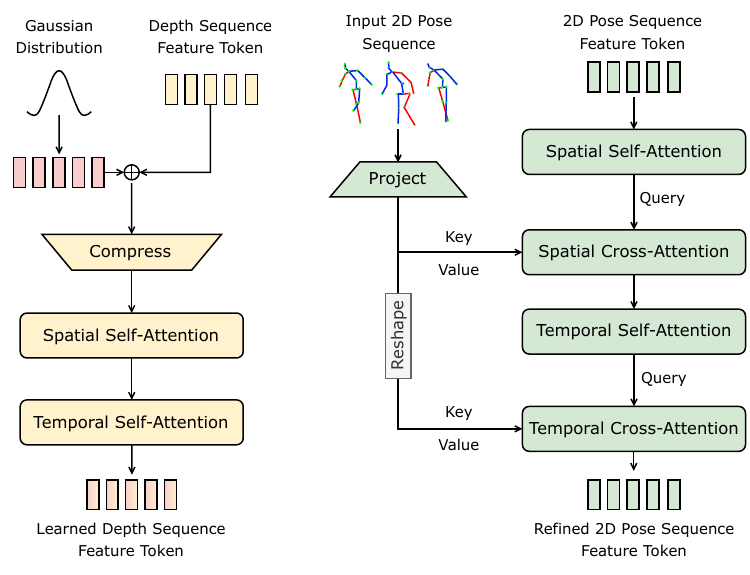}
\caption{Illustration of proposed 2D pose expert and depth expert. The 2D pose expert takes the 2D pose features and the 2D pose sequence as input, and outputs refined 2D pose features. The depth expert takes the depth features along with a supplementary feature initialized from a Gaussian distribution, and outputs the learned depth features. The supplementary feature is learnable during the training.
}
\label{fig:5}
\end{figure}

Specifically, the architecture of the 2D pose expert is illustrated in Figure~\ref{fig:5}. Since the input 2D pose and the 2D part of the output 3D pose share consistent high-level semantic information, requiring only a coordinate projection, we re-integrate the input 2D pose into the 2D pose expert to enhance the feature representation of the 2D pose feature. 
We use a linear layer to project the input 2D pose sequence to feature $F_{P} \in \mathbb{R}^{T \times J \times C}$. To enhance the 2D pose features, we first apply a multi-head self-attention~\cite{transformer} to $F_{2D}$ as Equation~\ref{eq:2dssa}. 
\begin{equation}
\label{eq:2dssa}
    F_{2D} = SpatialMHSA(F_{2D})
\end{equation}
Then, we take $F_{2D}$ as the query and $F_{P}$ as the key and value, and we perform multi-head cross-attention~\cite{transformer} to extract spatial prior information from the 2D pose as follows:
\begin{equation}
\label{eq:2dsca}
    F_{2D} = SpatialMHCA(F_{2D},F_{P}) 
\end{equation}
Based on the above process, the spatial structure of the 2D pose features is effectively strengthened. We then proceed to model the temporal relationships. Similarly, we first apply temporal multi-head self-attention to $F_{2D}$ as defined in Equation~\ref{eq:2dtsa}. 
\begin{equation}
\label{eq:2dtsa}
    F_{2D} = TemporalMHSA(F_{2D})
\end{equation}
Then, we perform multi-head cross-attention along the temporal dimension by using $F_{2D}$ as the query and $F_{P}$ as the key and value as follows:
\begin{equation}
\label{eq:2dtca}
    F_{2D} = TemporalCrossAttention(F_{2D},F_{P}) 
\end{equation}
Through the 2D pose expert, we iteratively refine the 2D pose feature representation using the input 2D pose, thereby enhancing the spatiotemporal correlation of the 2D pose features.

Then, we propose a depth expert to learn depth features. Although existing 2D-to-3D lifting methods achieve impressive performance, the depth information implicitly contained in the input 2D pose sequence remains limited. The inherent information entropy of the input 2D pose sequence limits the theoretical upper bound on performance. To address this limitation, we introduce a set of learnable tokens $F_{G} \in \mathbb{R}^{T \times J \times C}$ initialized from a Gaussian distribution as supplementary inputs. 
% The Gaussian distribution provides the depth expert with a statistical prior that more closely approximates its target features, which facilitates the model with the most natural constrained starting point, facilitating more stable convergence.
The Gaussian tokens are crucial because they serve as learnable tokens for the Depth Expert, providing a statistically justified prior (based on the Central Limit Theorem) that represents depth information for each joint, compensating for the limited information entropy of the 2D input and promoting stable convergence. Although existing 2D-to-3D lifting methods achieve impressive performance, the depth information implicitly contained in the input 2D pose sequence remains limited, as 2D causes visual representation reduction compared to raw images. The inherent information entropy of the input 2D pose sequence limits the theoretical upper bound on performance. Therefore, we introduce learnable tokens to supplement information during training. According to the central limit theorem states that, as the sample size increases, the distribution of a normalized version of the sample mean converges to a standard normal distribution. This holds even if the original variables themselves are not normally distributed. Consequently, our decision to initialize the learnable tokens with a Gaussian distribution strategically equates to providing the depth expert with a statistical prior that more closely approximates its target features—since the final, complex depth features are the accumulated result of numerous, independently varying factors (e.g., noise, 2D error, perspective effects). We further validated the effectiveness of this initialization by conducting ablation experiments comparing Gaussian initialization with zero initialization and Laplace initialization, which demonstrated that the Gaussian distribution achieves better performance.
\begin{equation}
\label{eq:gaussian}
q(F_{G}) = \mathcal{N}(F_{G};0,\mathbf{I})
\end{equation}
We use an MLP layer to compress and fuse the depth features with the Gaussian tokens as Equation~\ref{eq:compress}, introducing supplementary information while reducing computational overhead.
\begin{equation}
\label{eq:compress}
\begin{split}
    F_{D} = Compress(F_{D}\oplus F_{G})
\end{split}
\end{equation}
Subsequently, we apply spatial multi-head self-attention and temporal multi-head self-attention to the augmented depth features 
$F_{D}$ as follows:
% \begin{equation}
% \label{eq:expert}
%     F_{D} = SpatialMHSA(F_{D}) 
% \end{equation}
\begin{equation}
\label{eq:dest}
    F_{D} = TemporalMHSA(SpatialMHSA(F_{D})) 
\end{equation}

% \begin{equation}
%     F_{2D}  = \underbrace{Softmax(\mathcal{F}_{FC}(F_{2D} \oplus F_{D}))}_{\Theta} \left(\begin{array}{l}
% F_{2D} \\
% F_{D} \\
% \end{array}\right)
% = \theta_{1}F_{2D} + \theta_{2}F_{D}
% \end{equation}
Through the 2D pose expert and depth expert, both the 2D pose features and depth features have enhanced their feature representations. Subsequently, we introduce the Pose Router to perform a coarse-grained dynamic fusion of the outputs from the two experts. Both the 2D pose and depth features have their own gating layers as Equation~\ref{eq:2dgat} and~\ref{eq:degat}, which are responsible for determining the relevance of each feature, allowing for more efficient and accurate fusion.
\begin{equation}
\label{eq:2dgat}
    F_{2D}  = Softmax(\mathcal{F}^{2D}_{FC}(F_{2D} \oplus F_{D})) \left(\begin{array}{l}
F_{2D} \\
F_{D} \\
\end{array}\right)
\end{equation}
\begin{equation}
\label{eq:degat}
    F_{D}  = Softmax(\mathcal{F}^{D}_{FC}(F_{2D} \oplus F_{D})) \left(\begin{array}{l}
F_{2D} \\
F_{D} \\
\end{array}\right)
\end{equation}

\subsection{PoseMoE Decoder}

As shown in Figure~\ref{fig:6}, the PoseMoE Decoder comprises spatial and temporal cross-expert knowledge aggregation modules. Its primary objective is to model fine-grained global dependencies between 2D pose features and depth features before regressing the final 2D poses and depths. We first use the 2D pose features and depth features as the query and key-value to perform spatial multi-head cross-attention as follows:
\begin{equation}
\label{eq:dec2dca}
    F_{2D}, M_{2D \rightarrow D} = SpatialMHCA(F_{2D},F_{D}) 
\end{equation}
\begin{equation}
\label{eq:decdeca}
    F_{D}, M_{D \rightarrow 2D} = SpatialMHCA(F_{D},F_{2D}) 
\end{equation}
This operation enables the model to capture spatial dependencies between the 2D pose and depth features, facilitating the fusion of spatial information across both modalities. After the above process, we can obtain two cross-attention matrices $M_{2D \rightarrow D}$ and $M_{D \rightarrow 2D}$. We skillfully leverage them to further learn global correlation. Concretely, we use matrix multiplication to obtain an aggregation attention matrix as:
\begin{figure}[t] 
\centering
\includegraphics[width=1\linewidth]{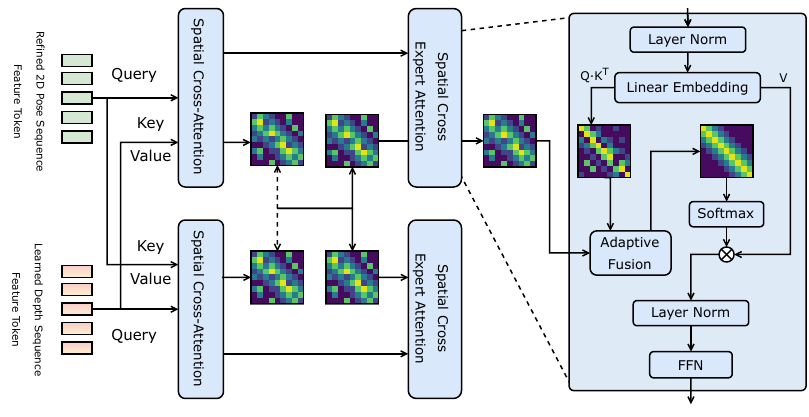}
\caption{The overview of Spatial Cross Expert Knowledge Aggregation includes both the 2D pose branch and the depth branch. Each branch contains spatial cross-attention and spatial cross-expert attention.
}
\label{fig:6}
\end{figure}
\begin{equation}
\label{eq8}
M_{2D \rightarrow 2D} = M_{2D \rightarrow D} \cdot M_{D \rightarrow 2D}
\end{equation}
\begin{equation}
\label{eq8}
M_{D \rightarrow D} = M_{D \rightarrow 2D} \cdot M_{2D \rightarrow D}
\end{equation}
Next, we use the spatial cross-expert attention in the 2D pose branch as an example to explain the process.
We map the 2D pose feature $F_{2D}$ to queries $Q_{F_{2D}}$, keys $K_{F_{2D}}$, and values $V_{F_{2D}}$ as follows: 
\begin{align}
    Q_{F_{2D}} &=F_{2D}\cdot W_{Q} \nonumber \\
    K_{F_{2D}} &=F_{2D}\cdot W_{K}  \\
    V_{F_{2D}} &=F_{2D}\cdot W_{V}  \nonumber 
\end{align}
The aggregation attention matrix is then adaptively fused with the original self attention matrix as follows:
\begin{align}
    \overline{M} =& Sigmoid(\mu)\cdot M_{2D \rightarrow 2D} \nonumber \\
    &+ (1 -Sigmoid(\mu) ) \cdot Q_{F_{2D}}\cdot K^{T}_{F_{2D}}
\end{align}
where $\mu$ is a learnable parameter for each layer during the training process to adaptively learn suitable fusion ratios.
Finally, we get the enhanced 2D pose sequence feature $F_{2D}$ as follows:
\begin{equation}
    F_{2D} = F_{2D} + Softmax(\overline{M}/\sqrt{C}) \cdot V_{F_{2D}} 
\end{equation}
The operations in the depth branch are similar to those in the 2D pose branch, with the difference being the use of the matrix  $M_{D \rightarrow D}$ instead of  $M_{2D \rightarrow 2D}$.
The model structure of Temporal Cross Expert Knowledge Aggregation is similar to Spatial Cross Expert Knowledge Aggregation, with the only difference being the transformation of the features' dimensional order.
\subsection{Regression Head and Loss Function}
\label{sub:towerandloss}
We use two regression heads (MLP) to regress the 2D pose $\overline{Y}_{2D} \in \mathbb{R}^{T \times J \times 2}$ and depth $\overline{Y}_{D}\in \mathbb{R}^{T \times J \times 1}$ respectively and concatenate them to generate the 3D pose sequence $\overline{Y}$. 
Losses are independently calculated for 2D pose and depth as Equation~\ref{eq:loss2d} and~\ref{eq:lossd}. For the 2D pose, we use L2 loss to minimize the errors between predictions and ground truth. For depth, we employ the mean absolute error loss to minimize the discrepancy between the estimated depth and the ground truth. Mathematically, this is equivalent to using the L2 distance, given that the depth has a dimension of 1.
\begin{align}
 \mathcal{L}_{2D} &= \color{black}\frac{1}{JT} \sum_{j=1}^{J} \sum_{t=1}^{T} \left\|Y_{2D}^{j,t}-\overline{Y}_{2D}^{j,t}\right\|_{2} \color{black} \label{eq:loss2d}\\
\mathcal{L}_{D} &=  \color{black}\frac{1}{JT} \sum_{j=1}^{J} \sum_{t=1}^{T}\left|Y_{D}^{j,t}-\overline{Y}_{D}^{j,t}\right| \color{black} \label{eq:lossd}
 % \mathcal{L}_{3D} &=  \mathcal{L}_{2D} +  \mathcal{L}_{D}
\end{align}
Where $Y_{2D}^{j,t}$ and $Y_{D}^{j,t}$ are the 2D pose and depth of 3D pose label.
$\overline{Y}_{2D}^{j,t}$ and $\overline{Y}_{D}^{j,t}$ are the predicted results of the $j$-th joint in $t$-th frame. In addition, the temporal consistency loss $\mathcal{L}_{T}$ from ~\cite{hossain2018exploiting} is introduced to produce smooth poses. 
The total loss $\mathcal{L}$ is defined as follows:
\begin{equation}
\mathcal{L} = \mathcal{L}_{2D}+\mathcal{L}_{D}+\lambda_{T}\mathcal{L}_{T}
\end{equation}
where $\lambda_{T}$ is hyper-parameter to balance the ratio of different loss terms.

\section{Experiments}
% \subsection{Experiment Setting}
% We evaluate our model on two large-scale monocular 3D human pose estimation datasets: Human3.6M~\cite{h36m} and MPI-INF-3DHP~\cite{3dhp}.
% %
% For the Human3.6M dataset, we report the MPJPE (Mean Per Joint Position Error) and P-MPJPE (Procrustes-MPJPE) as evaluation metrics as prior methods~\cite{mhformer,motionbert,mixste,poseformerv2}. 
% %
% For the MPI-INF-3DHP dataset, similar to existing approaches~\cite{pstmo,tang20233d,chen2023hdformer,motionbert}, we use ground truth 2D pose as input and report MPJPE, Percentage of Correct Keypoint (PCK) with the threshold of 150mm, and Area Under Curve (AUC) as the evaluation metrics. 
\subsection{Datasets and Evaluation Metrics}

\begin{table*}[t]
  \caption{Results on Human3.6M in millimeters (mm) under MPJPE using 2D pose detected by CPN~\cite{chen2018cascaded} following previous works~\cite{mhformer,poseformerv2,hot,peng2024ktpformer}. T is the length of the input 2D pose sequence. The best result is shown in bold, and the second-best result is underlined.}
  \label{tab:h36m}
  \centering
\scalebox{0.8}{
  \begin{tabular}{lc|c|ccccccccccccccc|c}
    \toprule
\textbf{MPJPE}&Venue &T& Dir. & Disc. & Eat & Greet & Phone & Photo & Pose & Pur. & Sit & SitD. & Smoke & Wait & WalkD. & Walk & WalkT. & Avg \\
    \midrule
MHFormer~\cite{mhformer} &CVPR'22& 81 &-&-&-&-&-&-&-&-&-&-&-&-&-&-&-&44.5 \\
MixSTE~\cite{mixste} &CVPR'22 & 81 &39.8& 43.0& 38.6& 40.1& 43.4& 50.6& 40.6& 41.4& 52.2& 56.7& 43.8& 40.8& 43.9& 29.4& 30.3& 42.4 \\
P-STMO~\cite{pstmo} &ECCV'22& 81 &41.7& 44.5& 41.0& 42.9& 46.0& 51.3& 42.8& 41.3& 54.9& 61.8& 45.1& 42.8& 43.8& 30.8& 30.7& 44.1 \\
PoseFormerV2~\cite{poseformerv2} &CVPR'23& 81&-&-&-&-&-&-&-&-&-&-&-&-&-&-&-&46.0 \\
STCFormer~\cite{tang20233d} &CVPR'23 & 81&40.6& 43.0& 38.3& 40.2& 43.5& 52.6& 40.3& 40.1& 51.8& 57.7& 42.8& 39.8& 42.3& 28.0& 29.5& 42.0 \\
GLA-GCN~\cite{yu2023gla} &ICCV'23& 81 &-&-&-&-&-&-&-&-&-&-&-&-&-&-&-&- \\
% MotionBERT~\cite{motionbert}& 81 &-&-&-&-&-&-&-&-&-&-&-&-&-&-&-&- \\
KTPFormer~\cite{peng2024ktpformer} &CVPR'24 &81 &39.1&41.9&37.3&40.1&44.0&51.3&39.8&41.0&51.4&56.0&43.0&41.0&42.6&28.8&29.5&41.8\\
PoseRetNet~\cite{zheng20243d} &ECCV'24 &81 &36.9 &40.5 &39.0 &38.6& 43.3& 49.6 &38.8 &40.2 &52.6 &56.5 & 42.6 &38.8& 40.5 &26.8 &28.4 &\underline{40.9} \\

\cellcolor{blue!5}\textbf{PoseMoE (Ours)}&\cellcolor{blue!5}-  & \cellcolor{blue!5}81  &\cellcolor{blue!5}39.7&\cellcolor{blue!5}41.4&\cellcolor{blue!5}39.4&\cellcolor{blue!5}35.5&\cellcolor{blue!5}43.1&\cellcolor{blue!5}50.7&\cellcolor{blue!5}40.0&\cellcolor{blue!5}37.2&\cellcolor{blue!5}51.1&\cellcolor{blue!5}56.0&\cellcolor{blue!5}43.7&\cellcolor{blue!5}40.5&\cellcolor{blue!5}39.1&\cellcolor{blue!5}28.7&\cellcolor{blue!5}28.8&\cellcolor{blue!5}\textbf{40.7} \\
\midrule
MHFormer~\cite{mhformer}&CVPR'22&351 & 39.2 & 43.1 & 40.1 & 40.9 & 44.9 & 51.2 & 40.6 & 41.3 & 53.5 & 60.3 & 43.7 & 41.1 & 43.8 & 29.8 & 30.6 & 43.0 \\
MixSTE~\cite{mixste}&CVPR'22 &243 &37.6& 40.9& 37.3& 39.7& 42.3& 49.9& 40.1& 39.8& 51.7& 55.0& 42.1& 39.8& 41.0& 27.9& 27.9& 40.9 \\
P-STMO~\cite{pstmo}&ECCV'22 & 243 &38.9& 42.7& 40.4& 41.1& 45.6& 49.7& 40.9& 39.9& 55.5& 59.4& 44.9& 42.2& 42.7& 29.4& 29.4& 42.8 \\
PoseFormerV2~\cite{poseformerv2}&CVPR'23 & 243 &-&-&-&-&-&-&-&-&-&-&-&-&-&-&-&45.2 \\
STCFormer~\cite{tang20233d}&CVPR'23 &  243&39.6& 41.6& 37.4& 38.8& 43.1& 51.1& 39.1& 39.7& 51.4& 57.4& 41.8& 38.5& 40.7& 27.1& 28.6& 41.0\\
GLA-GCN~\cite{yu2023gla}&ICCV'23 & 243 &41.3 & 44.3 & 40.8 & 41.8 & 45.9 & 54.1 & 42.1 & 41.5 & 57.8 & 62.9 & 45.0 & 42.8 & 45.9 & 29.4 & 29.9 & 44.4 \\
% MotionBERT~\cite{motionbert}&243 &\underline{36.3} & 38.7 & 38.6 & 33.6 &42.1 & 50.1 & \textbf{36.2} & \underline{35.7} & \underline{50.1} & 56.6 & 41.3 & \underline{37.4} & \underline{37.7} & \textbf{25.6} & \underline{26.5} & 39.2 \\
KTPFormer~\cite{peng2024ktpformer}&CVPR'24 &243 &37.3&39.2&35.9&37.6&42.5&48.2&38.6&39.0&51.4&55.9&41.6&39.0&40.0&27.0&27.4&\underline{40.1}\\
PoseRetNet~\cite{zheng20243d}&ECCV'24&243&36.9& 40.1 &38.7&38.3 &42.9&48.6&38.2&40.0&52.5 &55.4&42.3&38.7&39.7&26.2&27.8&
40.4\\

\cellcolor{blue!5}\textbf{PoseMoE (Ours)} &\cellcolor{blue!5}-  & \cellcolor{blue!5}243  &\cellcolor{blue!5}36.0&\cellcolor{blue!5}38.2&\cellcolor{blue!5}37.3&\cellcolor{blue!5}33.5&\cellcolor{blue!5}40.4&\cellcolor{blue!5}46.9&\cellcolor{blue!5}37.5&\cellcolor{blue!5}34.6&\cellcolor{blue!5}48.9&\cellcolor{blue!5}52.9&\cellcolor{blue!5}40.7&\cellcolor{blue!5}36.6&\cellcolor{blue!5}36.7&\cellcolor{blue!5}26.1&\cellcolor{blue!5}26.1&\cellcolor{blue!5}\textbf{38.7} \\
    \bottomrule
  \end{tabular}
  }
\end{table*}

\noindent\textbf{Human3.6M}~\cite{h36m} is the most popular benchmark for indoor 3D human pose estimation, which contains approximately 3.6 million frames captured by 4 cameras at different views. This dataset contains 11 subjects performing 15 typical actions (e.g., walking and sitting). To ensure a fair comparison, we follow previous methods~\cite{poseformer,mixste,tang20233d,motionbert} by using subjects 1, 5, 6, 7, and 8 for model training and subjects 9 and 11 for evaluation. 

\noindent\textbf{MPI-INF-3DHP}~\cite{3dhp} is a recently proposed large-scale challenging dataset with both indoor and outdoor scenes. The training set comprises 8 subjects, covering 8 activities, ranging from walking and sitting to complex exercise poses and dynamic actions. The test set covers 7 activities, containing three scenes: green screen, non-green screen, and outdoor environments. It complements existing test sets with more diverse motions (standing/walking, sitting/reclining, exercise, sports (dynamic poses), on the floor, dancing/miscellaneous).
\noindent\textbf{3DPW}~\cite{von2018recovering} is a challenging in-the-wild dataset, which contains several dynamic scenes. Also, it includes novel actions beyond the scope of Human3.6M, such as climbing and fencing. It includes training, test, and validation sets, consisting of 24, 24, and 12 videos, respectively. To verify the generalization of the proposed method to unseen scenarios, we follow the standard setting of previous works~\cite{von2018recovering,motionbert,li2024hyre} and use its test set for evaluation with MPJPE and PA-MPJPE as metrics.

\noindent\textbf{Evaluation Metrics.} For the Human3.6M~\cite{h36m} and 3DPW~\cite{von2018recovering} datasets, we use two common evaluation metrics: MPJPE and P-MPJPE. MPJPE (Mean Per Joint Position Error) is computed as the mean Euclidean distance between the estimated joints and the ground truth in millimeters after aligning their root joints (hip). P-MPJPE (Procrustes-MPJPE) is the MPJPE after the estimated joints align to the ground truth via a rigid transformation.
For the MPI-INF-3DHP~\cite{3dhp} dataset, following previous works~\cite{pstmo,tang20233d,chen2023hdformer,motionbert}, we use ground
truth 2D pose as input and report MPJPE, Percentage of
Correct Keypoint (PCK) with the threshold of 150mm, and Area Under Curve (AUC) as the evaluation metrics.
%
%
% Due to space limitations, we detail our experiment settings and implementations in Appendix~\ref{appendix:setting}.
% Please refer to Appendix~\ref{appendix:setting} for implementation details.
\subsection{Implementation details} 
Our model is implemented using PyTorch and executed on a server equipped with 2 NVIDIA 3090 GPUs. We apply horizontal flipping augmentation for both training and testing following~\cite{tang20233d,motionbert,foo2023unified,zhao2023contextaware}. For model training, we set each mini-batch as 16 sequences. The network parameters are optimized using AdamW~\cite{adamw} optimizer over 90 epochs with a weight decay of 0.01. The initial learning rate is set to 5e-4 with an exponential learning rate decay schedule and the decay factor is 0.99. 
% In the experiments on Human3.6M, two kinds of input are utilized, including the 2D ground truth and the Stacked Hourglass~\cite{newell2016stacked} 2D pose detection, following~\cite{motionbert,ci2019optimizing}. For MPI-INF-3DHP, 2D ground truth is used following previous works~\cite{poseformer,mixste,pstmo,motionbert}. While our proposed framework is capable of adapting to input sequences of any length, to be fair, we choose specific input sequence lengths (denoted as $T$) for two datasets to compare our method with other approaches that have a certain 2D input length~\cite{poseformer,mixste,pstmo,tang20233d}: Human3.6M ($T = 81, 243$), MPI-INF-3DHP ($T = 9, 27, 81$).

\begin{table*}[t]
  \caption{Results on Human3.6M in millimeters (mm) under P-MPJPE using 2D pose detected by CPN~\cite{chen2018cascaded} following previous works~\cite{mhformer,poseformerv2,hot,peng2024ktpformer}. T is the length of the input 2D pose sequence. The best result is shown in bold, and the second-best result is underlined.}
  \label{tab:h36mp2}
  \centering
\scalebox{0.8}{
  \begin{tabular}{lc|c|ccccccccccccccc|c}
    \toprule
\textbf{MPJPE}&Venue &T& Dir. & Disc. & Eat & Greet & Phone & Photo & Pose & Pur. & Sit & SitD. & Smoke & Wait & WalkD. & Walk & WalkT. & Avg \\
    \midrule
MHFormer~\cite{mhformer} &CVPR'22& 81 &-&-&-&-&-&-&-&-&-&-&-&-&-&-&-&- \\
MixSTE~\cite{mixste} &CVPR'22 & 81 &32.0& 34.2&31.7&33.7 &34.4 &39.2&32.0&31.8 &42.9& 46.9&35.5&
32.0&34.4&23.6&25.2& 33.9\\

P-STMO~\cite{pstmo} &ECCV'22& 81 &-&-&-&-&-&-&-&-&-&-&-&-&-&-&-&- \\
PoseFormerV2~\cite{poseformerv2} &CVPR'23& 81&-&-&-&-&-&-&-&-&-&-&-&-&-&-&-&- \\
STCFormer~\cite{tang20233d} &CVPR'23 & 81&30.4& 33.8&31.1&31.7&33.5&39.5&30.8&30.0& 41.8& 45.8&34.3&30.1&32.8&21.9&23.4&32.7\\

GLA-GCN~\cite{yu2023gla} &ICCV'23& 81 &-&-&-&-&-&-&-&-&-&-&-&-&-&-&-&- \\
% MotionBERT~\cite{motionbert}& 81 &-&-&-&-&-&-&-&-&-&-&-&-&-&-&-&- \\
KTPFormer~\cite{peng2024ktpformer} &CVPR'24 &81 &30.6&33.4 &30.1 &31.9&33.7& 38.2& 30.6& 30.7&40.9&44.8& 34.4&30.5&32.7&22.3&24.0&\textbf{32.6}\\
PoseRetNet~\cite{zheng20243d} &ECCV'24 &81 &30.5& 33.1& 31.4 &31.6& 33.0&38.4& 29.8& 30.6 &43.6 &45.4& 34.4& 30.3&32.4&21.5& 22.2&\textbf{32.6}\\

\cellcolor{blue!5}\textbf{PoseMoE (Ours)}&\cellcolor{blue!5}-  & \cellcolor{blue!5}81  &\cellcolor{blue!5}31.1&\cellcolor{blue!5}32.7&\cellcolor{blue!5}31.2&\cellcolor{blue!5}30.6&\cellcolor{blue!5}33.0&\cellcolor{blue!5}38.1&\cellcolor{blue!5}29.7&\cellcolor{blue!5}31.3&\cellcolor{blue!5}41.8&\cellcolor{blue!5}47.3&\cellcolor{blue!5}34.3&\cellcolor{blue!5}30.5&\cellcolor{blue!5}32.7&\cellcolor{blue!5}22.0&\cellcolor{blue!5}23.6&\cellcolor{blue!5}\textbf{32.6} \\
\midrule
MHFormer~\cite{mhformer}&CVPR'22&351 & 31.5 &34.9&32.8&33.6&35.3&39.6&32.0&32.2 &43.5&48.7&36.4&32.6&34.3&23.9&25.1&34.4\\
MixSTE~\cite{mixste}&CVPR'22 &243 &32.0&34.2&31.7&33.7&34.4&39.2&32.0&31.8&42.9&46.9&35.5&32.0&34.4&23.6&25.2&33.9\\
P-STMO~\cite{pstmo}&ECCV'22 & 243 &31.3&35.2&32.9&33.9& 35.4&39.3&32.5& 31.5& 44.6& 48.2&36.3&32.9&34.4&23.8&23.9&34.4\\
PoseFormerV2~\cite{poseformerv2}&CVPR'23 & 243 &-&-&-&-&-&-&-&-&-&-&-&-&-&-&-&35.6 \\
STCFormer~\cite{tang20233d}&CVPR'23 &  243&29.3&33.0& 30.7&30.6&32.7&38.2& 29.7&28.8& 42.2& 45.0&33.3&29.4&31.5&20.9&22.3&\underline{31.8}\\
GLA-GCN~\cite{yu2023gla}&ICCV'23 & 243 &32.4&35.3&32.6&34.2&35.0&42.1&32.1& 31.9& 45.5&49.5&36.1& 32.4 &35.6& 23.5&24.7& 34.8\\
KTPFormer~\cite{peng2024ktpformer}&CVPR'24 &243 &30.1&32.3&29.6&30.8&32.3&37.3&
30.0&30.2&41.0&45.3&33.6&29.9&31.4&21.5&22.6&31.9\\
PoseRetNet~\cite{zheng20243d}&ECCV'24&243&30.8 &33.1& 31.3& 31.8 &33.4 &37.7& 30.1 &30.5& 43.4& 45.5 &34.3& 30.3& 31.5 &21.4 &22.7&32.5\\
\cellcolor{blue!5}\textbf{PoseMoE (Ours)} &\cellcolor{blue!5}-  & \cellcolor{blue!5}243  &\cellcolor{blue!5}29.3&\cellcolor{blue!5}31.7&\cellcolor{blue!5}30.5&\cellcolor{blue!5}29.0&\cellcolor{blue!5}31.6&\cellcolor{blue!5}36.4&\cellcolor{blue!5}27.8&\cellcolor{blue!5}29.4&\cellcolor{blue!5}40.0&\cellcolor{blue!5}43.6&\cellcolor{blue!5}32.9&\cellcolor{blue!5}28.8&\cellcolor{blue!5}21.3&\cellcolor{blue!5}30.7&\cellcolor{blue!5}22.1&\cellcolor{blue!5}\textbf{31.0} \\
    \bottomrule
  \end{tabular}
  }
\end{table*}

\begin{table*}[t]
  \caption{Results on Human3.6M in millimeters (mm) under MPJPE using ground truth 2D pose. T is the number of input frames. Seq2seq refers to estimating 3D pose sequences rather than only the center frame. MACs/frames represents multiply-accumulate operations for each output frame. The best result is shown in bold, and the second-best result is underlined.}
  \label{tab:h36mgt}
  \centering

  \begin{tabular}{lc|ccc|cccc}
    \toprule
    Method &Venue    & Network     & Seq2Seq& T &Parameter&MACs &MACs/frame &MPJPE \\
    \midrule

MHFormer~\cite{mhformer}& CVPR'22& Lifting-Based &×&351& 30.9M&7.1G &7096M &30.5\\
MixSTE~\cite{mixste} &CVPR'22& Lifting-Based &\checkmark&243& 33.6M& 139.0G&572M & 21.6\\
P-STMO~\cite{pstmo} &ECCV'22& Lifting-Based &×&243& 6.2M&0.7G &740M & 29.3\\
PoseFormerV2~\cite{poseformerv2}&CVPR'23&Lifting-Based &×&243& 14.3M&0.5G &528M & -\\
STCFormer~\cite{tang20233d}& CVPR'23& Lifting-Based &\checkmark&243& 4.7M&19.6G & 80M& 21.3\\
GLA-GCN~\cite{yu2023gla}& ICCV'23& Lifting-Based &×&243&1.3M &1.5G &1556M & 21.0\\
% MotionBERT~\cite{motionbert} &ICCV'23& Lifting-Based &\checkmark&243&42.5M & 174.7G& 719M& 17.8\\
KTPFormer~\cite{peng2024ktpformer}& CVPR'24& Lifting-Based &\checkmark&243&33.7M&69.5G &286M & \underline{19.0}\\
PoseRetNet~\cite{zheng20243d}& ECCV'24& Lifting-Based &\checkmark&243&25.2M&104.5G &430M & 21.5\\
% \textcolor{magenta}{MotionAGFormer}~\cite{motionagformer2024}& WACV'24& Lifting-Based &\checkmark&243&19.0M&78.3G &322M & \underline{17.3}\\
\cellcolor{blue!5}\textbf{PoseMoE (Ours)}&\cellcolor{blue!5}-&\cellcolor{blue!5}Mixture-of-Experts&\cellcolor{blue!5}\checkmark&\cellcolor{blue!5}243&\cellcolor{blue!5}17.9M&\cellcolor{blue!5}66.3G&\cellcolor{blue!5}272M&\cellcolor{blue!5}\textbf{16.5}\\
    \bottomrule
  \end{tabular}

\end{table*}
\subsection{Comparison with State-of-the-art Methods}
\label{sec:sota}

\noindent\textbf{Human3.6M.} We compare our method with several state-of-the-art techniques on the Human3.6M dataset. 
For fair comparisons, only the results of models without extra pre-training on additional data are included. 
Table~\ref{tab:h36m} summarizes the performance comparisons in terms of MPJPE of all 15 actions, and the number of the input frames T is also given for each method. 
Our method achieved state-of-the-art performance with an MPJPE of 38.7mm with T = 243. It is worth noting that our method in the case of T = 81 input frames still achieves state-of-the-art performance with an MPJPE error of 40.7mm and even surpasses the performance of several methods with a higher number of input frames. 
For example, this result outperforms P-STMO~\cite{pstmo} (40.7mm v.s. 42.8mm), PoseformerV2~\cite{poseformerv2} (40.7mm v.s. 45.2mm) with 243 frames, and MHFormer~\cite{mhformer} even with 351 frames (40.7mm v.s. 43.0mm). 
These results demonstrate the effectiveness of PoseMoE. 
To further validate the effectiveness of the multi-task learning framework, we also report the model parameters,
MACs (Multiply–Accumulate Operations), and MPJPE using 2D ground truth as input. 
As shown in Table~\ref{tab:h36mgt}, our method with T = 243 achieves the best performance with an MPJPE of 16.5mm, which outperforms the lifting-based framework with faster inference speed. 
For example, this result outperforms KTPFormer~\cite{peng2024ktpformer} (16.5mm v.s. 19.0mm).

\noindent\textbf{MPI-INF-3DHP.} To demonstrate the generalization capability of our model, we also evaluate our model on the challenging MPI-INF-3DHP dataset, which includes more complex scenes and motions. 
Following previous works~\cite{poseformer,mixste,pstmo,tang20233d,mhformer,hot}, we use ground truth 2D pose as input and set the number of input frames as 9, 27, or 81.
As observed in Table~\ref{tab:mpi}, our method with T = 81 achieves the best performance with the PCK of 99.1\%, AUC of 86.9\%, and MPJPE of 16.3mm. 
Similar to the previous findings, our method with T = 9, 27 input frames still outperforms the previous state-of-the-art methods and achieves the MPJPE of 21.6mm and 18.5mm, respectively.
More remarkably, our method with T = 9 input frames outperforms the GLA-GCN~\cite{yu2023gla} with T = 81 input frames, despite having only one-ninth of the input frames (21.6mm v.s. 27.8mm, 9 frames vs. 81 frames).

\begin{table}[h]

  \caption{Results on MPI-INF-3DHP dataset under PCK, AUC, and MPJPE using ground truth 2D pose as input. T is the number of input frames. Seq2seq refers to estimating 3D pose sequence. }
  \label{tab:mpi}
  \centering
  \resizebox{1\columnwidth}{!}{
  \begin{tabular}{lc|c|c|ccc}
    \toprule
    Method  &Venue   & T   & Seq2Seq &PCK$\uparrow$&AUC$\uparrow$&MPJPE$\downarrow$\\
    \midrule
% PoseFormer~\cite{poseformer} &ICCV'21     & 9 &× & 88.6&56.4&77.1\\
MHFormer~\cite{mhformer} &CVPR'22   & 9 &× & 93.8&63.3&58.0\\
% STCFormer~\cite{tang20233d}    & 9 &\checkmark & \textbf{98.2}&\underline{81.5}&\underline{28.2}\\
% \midrule
MixSTE~\cite{mixste} &CVPR'22   & 27 &\checkmark & 94.4&66.5&54.9\\
% GLA-GCN~\cite{yu2023gla}    & 27 &× & 98.2&76.5&31.4\\
% STCFormer~\cite{tang20233d}      & 27 &\checkmark & \underline{98.4}&\underline{83.4}&\underline{24.2}\\
% KTPFormer~\cite{peng2024ktpformer} &27 &\checkmark &\textbf{98.9} &\underline{84.4}&19.2\\
% \midrule
P-STMO~\cite{pstmo}  &ECCV'22    & 81 &× & 97.9&75.8&32.2\\
PoseFormerV2~\cite{poseformerv2}  &CVPR'23    & 81&×  & 97.9&78.8&27.8\\
GLA-GCN~\cite{yu2023gla}  &ICCV'23    & 81 &× & 98.5&79.1&27.8\\
STCFormer~\cite{tang20233d}  &CVPR'23   & 81 &\checkmark & 98.7&83.9&23.1\\
% MotionBERT*~\cite{motionbert} &81 &\checkmark &\underline{98.7}&85.6& 16.5\\
KTPFormer~\cite{peng2024ktpformer}&CVPR'24 &81 &\checkmark &\underline{98.9} &\underline{85.9}&\underline{16.7}\\
% \textcolor{magenta}{MotionAGFormer}~\cite{motionagformer2024} &81 &\checkmark & 98.2 & 85.3&\underline{16.2}\\
PoseRetNet~\cite{zheng20243d}&ECCV'24 &81 &\checkmark &\textbf{99.1} &84.4&22.2\\
\cellcolor{blue!5}\textbf{PoseMoE} (Ours)&\cellcolor{blue!5}- & \cellcolor{blue!5}9&\cellcolor{blue!5}\checkmark&\cellcolor{blue!5}98.1&\cellcolor{blue!5}82.2&\cellcolor{blue!5}21.6\\
\cellcolor{blue!5}\textbf{PoseMoE} (Ours)  &\cellcolor{blue!5}-&\cellcolor{blue!5}27&\cellcolor{blue!5}\checkmark&\cellcolor{blue!5}98.6&\cellcolor{blue!5}85.8&\cellcolor{blue!5}18.5\\
\cellcolor{blue!5}\textbf{PoseMoE} (Ours) &\cellcolor{blue!5}-&\cellcolor{blue!5}81&\cellcolor{blue!5}\checkmark&\cellcolor{blue!5}\textbf{99.1}&\cellcolor{blue!5}\textbf{86.9}&\cellcolor{blue!5}\textbf{16.3}\\
    \bottomrule
  \end{tabular}
  }
\end{table}

\noindent\textbf{3DPW}. Table~\ref{tab:3dpw} presents the quantitative results of our proposed PoseMoE on the challenging 3DPW~\cite{von2018recovering} dataset, evaluated under both MPJPE and P-MPJPE metrics. Notably, PoseMoE consistently achieves state-of-the-art performance across all compared methods. Our method yields an MPJPE of 76.8mm and a P-MPJPE of 50.6mm. Specifically, PoseMoE significantly outperforms concurrently leading approaches in MPJPE, such as KTPFormer~\cite{peng2024ktpformer} and PoseRetNet~\cite{zheng20243d}. This demonstrates PoseMoE's superior capability in accurately estimating 3D poses. For the P-MPJPE metric, PoseMoE also exhibits a strong performance, achieving 50.6mm, surpassing KTPFormer~\cite{peng2024ktpformer} (52.7mm) and PoseRetNet~\cite{zheng20243d} (61.0mm). This further validates the robustness of our method in pose estimation after rigid alignment. Compared to earlier methods such as MHFormer~\cite{mhformer} and PoseFormerV2~\cite{poseformerv2}, PoseMoE shows substantial improvements in both metrics, highlighting the advancements of our approach in handling complex pose estimation tasks. In summary, across both MPJPE and P-MPJPE, PoseMoE not only surpasses its closest competitor but also demonstrates a consistent performance increase over representative works~\cite{mhformer,mixste,pstmo,poseformerv2,tang20233d,yu2023gla,peng2024ktpformer,zheng20243d}. These comprehensive results show the effectiveness and competitiveness of our proposed PoseMoE.

\begin{table}[h]

  \caption{Results on 3DPW dataset under MPJPE and P-MPJPE. The best result is shown in bold, and the second-best result is underlined.  }
  \label{tab:3dpw}
  \centering
  % \resizebox{1\columnwidth}{!}{
  \begin{tabular}{l|c|cc}
    \toprule
    Method  &  Venue   &MPJPE$\downarrow$&P-MPJPE$\downarrow$\\
    \midrule
% VideoPose~\cite{videopose}& CVPR'19&101.8&63.0\\
% PoseFormer~\cite{poseformer}& ICCV'21&118.2&73.1\\
MHFormer~\cite{mhformer} &CVPR'22&92.3&56.7\\
MixSTE~\cite{mixste} &CVPR'22&87.4&54.3 \\
% PCT~\cite{geng2023human} &CVPR'23&83.1&53.9\\
P-STMO~\cite{pstmo}&CVPR'22&97.6&63.7 \\
PoseFormerV2~\cite{poseformerv2}&CVPR'23&91.1&56.2 \\
STCFormer~\cite{tang20233d}&CVPR'23&84.9&54.5 \\
GLA-GCN~\cite{yu2023gla}&ICCV'23&91.6&55.8 \\
KTPFormer~\cite{peng2024ktpformer}& CVPR'24&\underline{81.0}&\underline{52.7}\\
PoseRetNet~\cite{zheng20243d}& ECCV'24&94.6&61.0\\
\cellcolor{blue!5}\textbf{PoseMoE (Ours)} &\cellcolor{blue!5}-&\cellcolor{blue!5}\textbf{76.8}&\cellcolor{blue!5}\textbf{50.6}\\
    \bottomrule
  \end{tabular}
  % }
\end{table}

\begin{figure}[h] 
\centering
\includegraphics[width=1\linewidth]{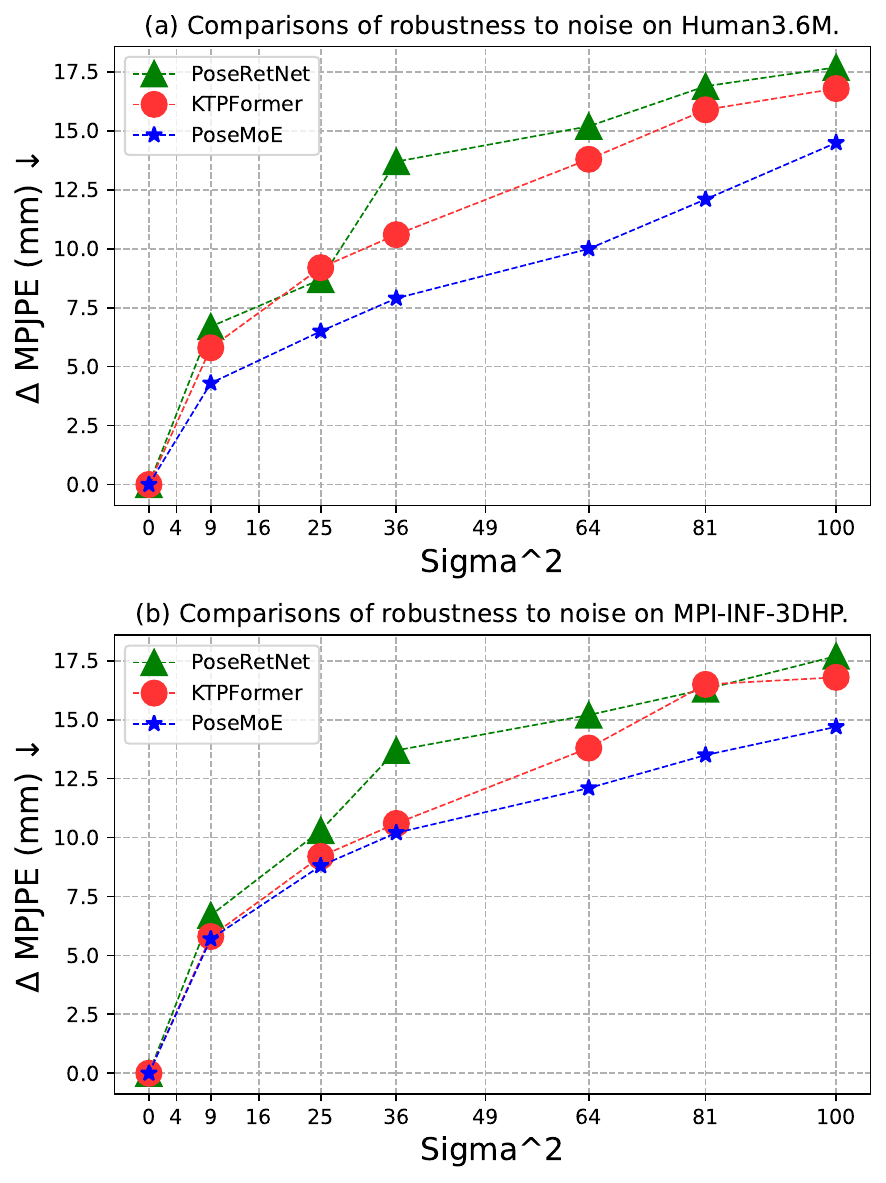}
\caption{Comparisons of PoseMoE, KTPFormer~\cite{peng2024ktpformer} and PoseRetNet~\cite{zheng20243d} in terms of robustness to noise on Human3.6M and MPI-INF-3DHP datasets. Zero-mean Gaussian noise of standard deviation sigma is added to ground truth 2D pose, and we show their performance drop ( $\Delta$ MPJPE, in millimeters) as sigma increases. The size of markers indicates the computational cost of models.
}
\label{fig:noise}
\end{figure}

\noindent\textbf{Robustness to Noisy 2D Pose.} Benefiting from the 2D pose branch, our method not only preserves well-detected 2D pose features but also allows us to handle noisy 2D pose input. To demonstrate that the inclusion of the 2D pose branch helps improve the robustness of the proposed method, we make the pose estimation task more challenging by adding zero-mean Gaussian noise to the ground-truth 2D pose on the Human3.6M~\cite{h36m} and MPI-INF-3DHP~\cite{3dhp}. As shown in Figure~\ref{fig:noise}, the experimental evidence reveals that our proposed PoseMoE suffers from less performance drop as the standard deviation of Gaussian noise (sigma) increases compared with the powerful lifting-based method KTPFormer~\cite{peng2024ktpformer} while being more efficient.

\subsection{Ablation Study}
We perform extensive ablation studies focused on analyzing the contribution of each component in our proposed PoseMoE. Experiments are conducted on the Human3.6M~\cite{h36m} dataset with T = 243 as the number of input frames and MPJPE is used as the evaluation metric.

\noindent\textbf{Analysis on Effectiveness of Components.} 
\begin{table}[t]
  \caption{Analysis of  each component within PoseMoE.}
  \label{tab:component}
  \centering
  \begin{tabular}{l|cc}
    \toprule
Model Setting & MPJPE&P-MPJPE\\
\midrule
Multi-Task Learning Baseline&42.1&33.6\\
 + PME (w/o 2D Expert)&40.6&32.5\\
 + PME (w/o Depth Expert)&41.3&32.7\\
 + PME (w/o Pose Router)&40.2&32.0\\
 + PME&39.6&31.8\\
 \midrule
 + PMD (w/o SpatialCEKA)&41.2&32.9\\
 + PMD (w/o TemporalCEKA)&41.5&33.1\\
 + PMD&40.7&32.5\\
 \midrule
 + PME + PMD &38.7&31.0\\
    \bottomrule
  \end{tabular}
\end{table}
To evaluate the effectiveness of each component in our proposed PoseMoE framework, we conduct a comprehensive ablation study, as shown in Table~\ref{tab:component}. We begin with a baseline model and progressively incorporate different modules of PoseMoE to assess their individual contributions. Our baseline configuration: a pure dual-branch network that adopts the MixSTE encoder. The parameter count of this baseline is roughly equivalent to that of the final PoseMoE to ensure experimental fairness. As shown in the table, simply replacing the identical, non-specialized encoder in the dual-branch baseline with our proposed PoseMoE Encoder, which incorporates modules specifically designed for the specialized refinement of 2D features and the learning of depth features, yields a substantial 2.5 mm improvement in the MPJPE. This significant empirical gain serves as powerful evidence validating our core theoretical premise: the structural decoupling afforded by the Mixture-of-Experts architecture is essential for mitigating feature entanglement and the detrimental effect of depth uncertainty. Furthermore, experimental results also clearly demonstrate the individual effectiveness of the subsequent fusion stage, showing that incorporating the PoseMoE Decoder into the established dual-branch network achieves a measurable 1.4 mm improvement on its own, thereby verifying the efficacy and independent contribution of our proposed strategies and mechanism for delayed, selective knowledge fusion. When utilizing the complete PoseMoE architecture, which features the structurally superior PoseMoE Encoder combined with the strategically optimized PoseMoE Decoder, our method achieves a remarkable 3.4 mm overall improvement compared to the dual-branch baseline, all while meticulously maintaining a roughly equivalent parameter count. This cumulative evidence establishes that the novelty and efficacy of PoseMoE reside not in model size, but in its principled architectural design that strategically decouples and then selectively aggregates the features based on their intrinsic difficulty and reliability.
% We first evaluate the impact of each submodule within the PoseMoE Encoder (PME). Removing either the 2D Pose Expert or the Depth Expert significantly degrades performance, indicating that both experts are essential for capturing complementary information from the input modalities. Furthermore, excluding the Pose Router results in noticeable performance drops, demonstrating its role in effectively fusing the outputs from the two experts through dynamic gating.
% Next, we analyze the PoseMoE Decoder (PMD) by disabling key components within the Cross Expert Knowledge Aggregation (CEKA). The removal of Spatial CEKA or Temporal CEKA leads to performance deterioration, confirming the importance of modeling fine-grained spatial and temporal dependencies. Integrating both PME and PMD yields the best performance, validating the overall effectiveness of the proposed PoseMoE architecture.

\begin{table}[t]
  \caption{Analysis for micro design within 2D pose expert.}
  \label{tab:2dexpert}
  \centering
  \begin{tabular}{l|cc}
    \toprule
Model Setting & MPJPE&P-MPJPE\\
\midrule
2D Pose Expert&38.7&31.0\\
\quad w/o SptialMHSA&39.4&31.7\\
 \quad w/o TemporalMHSA&39.6&32.0\\
 \quad w/o SptialMHCA&39.0&31.5\\
\quad  w/o TemporalMHCA&39.2&31.5\\
    \bottomrule
  \end{tabular}
\end{table}
\begin{table}[t]
  \caption{Analysis for micro design within depth expert.}
  \label{tab:depthexpert}
  \centering
  \begin{tabular}{l|cc}
    \toprule
Model Setting & MPJPE&P-MPJPE\\
\midrule
Depth Expert&38.7&31.0\\
\quad w/o Gaussion Token&39.2&31.4\\
\quad  w/o SpatialMHSA&39.5&31.6\\
 \quad w/o TemporalMHSA&39.6&31.9\\
    \bottomrule
  \end{tabular}
\end{table}

\begin{table}[h]
  \caption{Analysis for initialization distribution within depth expert.}
  \label{tab:initialization}
  \centering
  \begin{tabular}{l|cc}
    \toprule
Initialization & MPJPE&P-MPJPE\\
\midrule
Laplace Initialization&39.8&32.4\\
Zero Initialization&41.5&34.3\\
Gaussion Initialization&38.7&31.0\\
    \bottomrule
  \end{tabular}
\end{table}

\noindent\textbf{Analysis on PoseMoE Encoder.} 
To better understand the internal design choices of the 2D Pose Expert, we conduct a detailed ablation study on its key components, as shown in Table~\ref{tab:2dexpert}. Specifically, we analyze the impact of each attention mechanism, including spatial and temporal multi-head self-attention (MHSA), as well as spatial and temporal multi-head cross-attention (MHCA).
Removing the spatial MHSA leads to a notable performance drop, indicating that modeling spatial dependencies among joints is critical for accurate 2D pose feature representation. Similarly, excluding the temporal MHSA impairs the model’s ability to capture motion patterns over time, which are essential for sequential pose estimation.
Furthermore, the spatial MHCA module plays a key role in integrating supplementary information from the input 2D pose, and its removal results in a clear decline in accuracy. Likewise, the absence of temporal MHCA limits the network’s capacity to incorporate temporal priors, leading to degraded performance.
These results collectively demonstrate the necessity of each attention mechanism within the 2D Pose Expert, and confirm the effectiveness of our carefully designed spatiotemporal architecture.

To further validate the effectiveness of the internal design components in the Depth Expert, we conduct ablation experiments as summarized in Table~\ref{tab:depthexpert}. We investigate three key elements: the Gaussian tokens used as supplementary inputs, spatial multi-head self-attention (MHSA), and temporal MHSA.
Removing the Gaussian tokens leads to a significant drop in performance, which confirms their role in providing learnable priors that compensate for the lack of depth information and enrich the depth feature representation.
Moreover, the spatial MHSA is essential for capturing local spatial relationships among joints in a single frame. Its removal impairs the model's ability to reason about per-frame depth cues, resulting in higher errors. Similarly, the absence of temporal MHSA degrades the model's ability to exploit temporal consistency in joint depth across frames, which is crucial for stable and accurate depth estimation in motion sequences.
\begin{table}[t]
  \caption{Analysis for micro design within cross expert knowledge attention.}
  \label{tab:ceka}
  \centering
  \begin{tabular}{l|cc}
    \toprule
Model Setting & MPJPE&P-MPJPE\\
\midrule
SpatialCEKA&38.7&31.0\\
\quad w/o Cross Attention&39.1&31.6\\
 \quad w/o Cross Expert Attention&39.3&31.2\\
 \midrule
TemporalCEKA&38.7&31.0\\
\quad w/o Cross Attention&39.6&31.7\\
 \quad w/o Cross Expert Attention&39.5&31.5\\
    \bottomrule
  \end{tabular}
\end{table}
\begin{table}[t]
  \caption{Analysis for various lifting methods trained with our proposed PoseMoE Decoder.}
  \label{tab:pmd}
  \centering
  \begin{tabular}{lc|cc}
    \toprule
Model Setting &Venue& MPJPE&P-MPJPE\\
\midrule
MixSTE~\cite{mixste}& CVPR22&40.9&32.6\\
MixSTE~\cite{mixste} + PMD&-&39.8&31.9\\
\midrule
KTPFormer~\cite{peng2024ktpformer}&CVPR24&40.1&31.9\\
KTPFormer~\cite{peng2024ktpformer} + PMD&-&39.5&31.6\\
\midrule
PoseRetNet~\cite{zheng20243d}&ECCV24&40.4&32.5\\
PoseRetNet~\cite{zheng20243d} + PMD&-&40.1&32.0\\
    \bottomrule
  \end{tabular}
\end{table}

These results clearly demonstrate that each component within the Depth Expert contributes significantly to the overall performance, and highlight the importance of both spatial and temporal modeling in depth feature learning. 
To further validate the effectiveness of Gaussian tokens, we conducted ablation experiments comparing Gaussian initialization with zero initialization and Laplace initialization. As shown in Table~\ref{tab:initialization}, the Gaussian initialization achieves the best performance. This confirms that Gaussian initialization provides better prior knowledge initialization for the Depth Expert, thereby enhancing the model’s depth estimation capability.

\begin{figure*}[t] 
\centering
\includegraphics[width=1\linewidth]{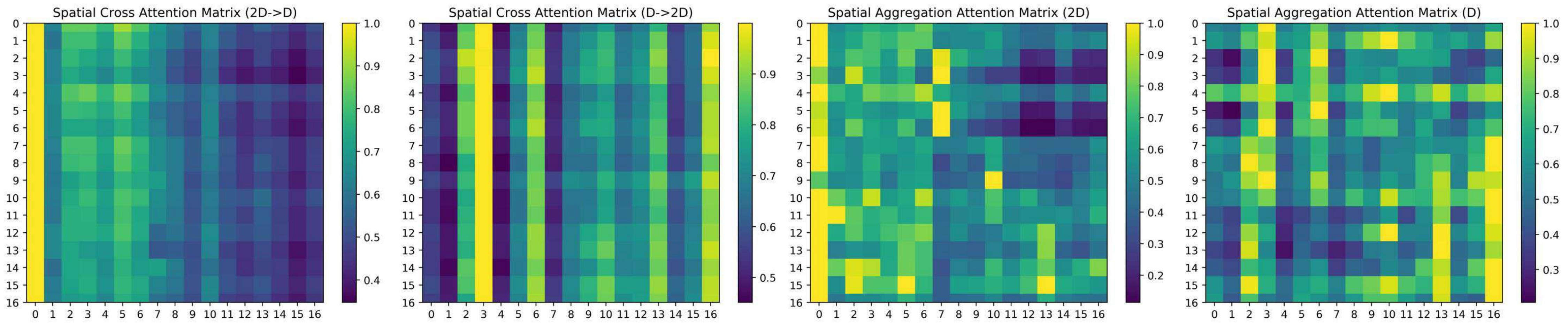}
\caption{Visualizations of different spatial attention matrices within PoseMoE Decoder. Through the bidirectional cross-attention mechanism between 2D and depth features, the learning of relationships among spatial joints becomes more comprehensive.
}
\label{fig:11}
\end{figure*}
\begin{figure*}[t] 
\centering
\includegraphics[width=1\linewidth]{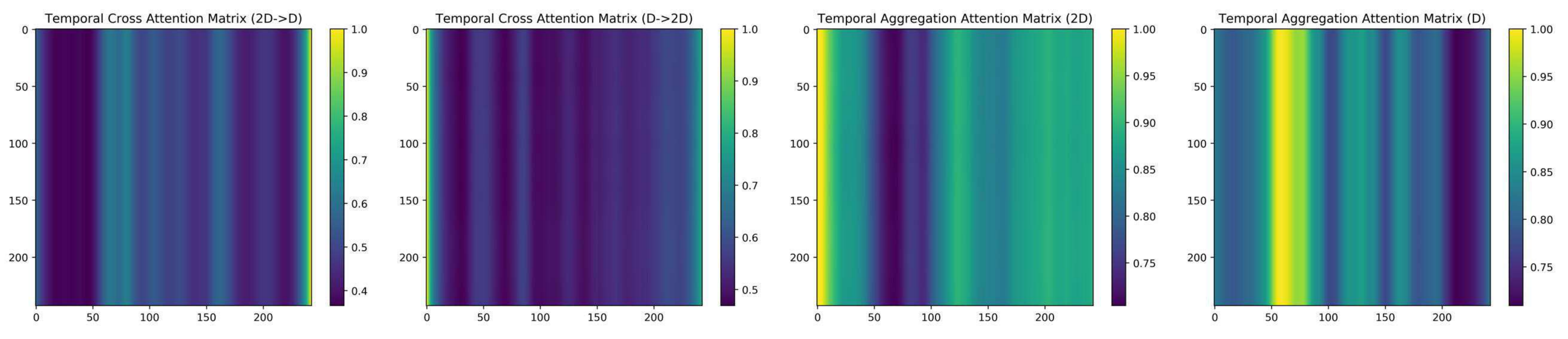}
\caption{Visualizations of different temporal attention matrices within PoseMoE Decoder. Through the bidirectional cross-attention mechanism between 2D and depth features, the learning of relationships among temporal motion becomes more comprehensive.
}
\label{fig:12}
\end{figure*}

\noindent\textbf{Analysis on PoseMoE Decoder.} We analyze the impact of key components in our Cross Expert Knowledge Attention (CEKA) module, as shown in Table~\ref{tab:ceka}. For both SpatialCEKA and TemporalCEKA variants, removing the cross-attention mechanism ("w/o Cross Attention") or disabling cross-expert interaction ("w/o Cross Expert Attention") leads to performance degradation in both MPJPE and P-MPJPE metrics. This highlights the necessity of integrating cross-domain knowledge and expert-specific attention to achieve optimal pose estimation accuracy. We also evaluate our proposed PoseMoE Decoder (PMD) by integrating it with state-of-the-art lifting methods: MixSTE~\cite{mixste}, KTPFormer~\cite{peng2024ktpformer}, and PoseRetNet~\cite{zheng20243d} in Table~\ref{tab:pmd}. Results demonstrate consistent improvements in MPJPE and P-MPJPE across all baselines when augmented with PMD. 
\begin{figure}[h]
\centering
\includegraphics[width=1\linewidth]{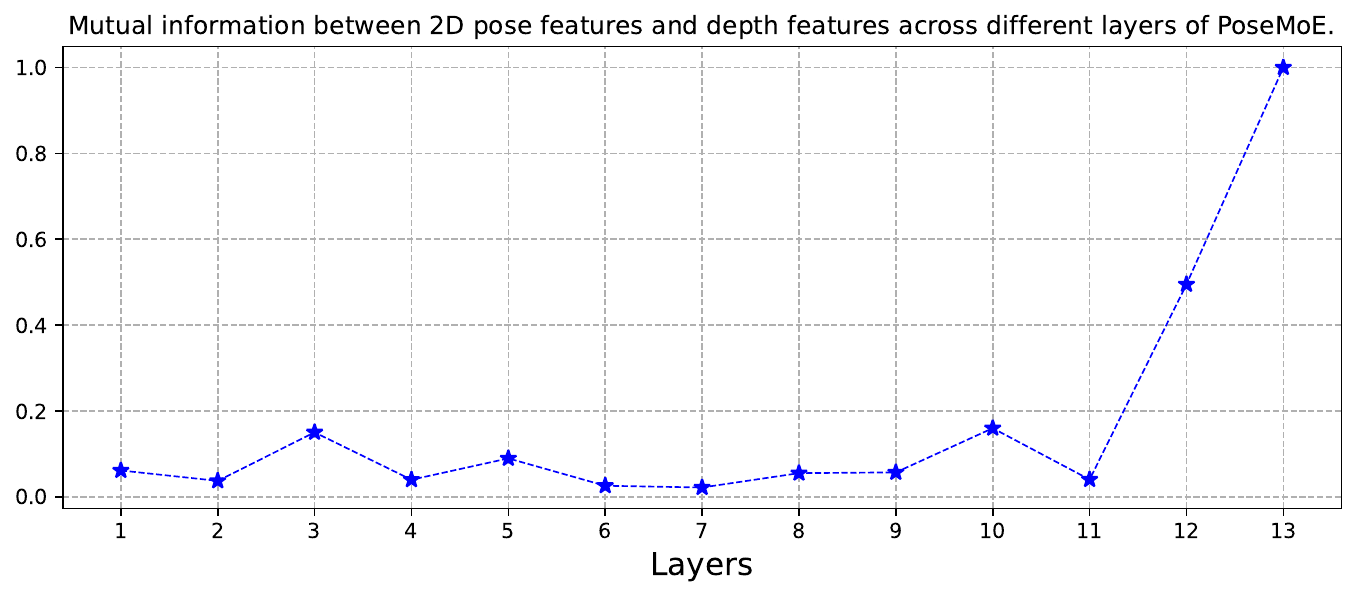}
\caption{Mutual information between 2D pose features and depth features across different layers of PoseMoE.The first 12 layers correspond to the PoseMoE Encoder, and the 13th layer corresponds to the PoseMoE Decoder. We normalize the values to the range [0, 1], using the values of the 13th layer as the reference standard.
}
\label{fig:13}
\end{figure}

\noindent\textbf{Analysis on Mutual Information within PoseMoE.} We calculated the mutual information between the 2D pose features and the depth features across different layers of our PoseMoE to provide robust theoretical support for our "Decoupling then Aggregation" design strategy. As shown in Figure~\ref{fig:13}, PoseMoE achieves low MI structural decoupling via the PoseMoE Encoder to solve the erosion problem, and then achieves high MI strategic aggregation via the PoseMoE Decoder to maximize the value of knowledge complementarity. The MI curve shows the relationship between 2D pose and depth features across layers, where Layers 1-12 are the PoseMoE Encoder and Layer 13 is the PoseMoE Decoder. The MI remains consistently low (well below 0.2) throughout the Encoder layers (1-12). This low MI confirms that the specialized experts successfully achieve active feature decoupling, preventing depth uncertainty from eroding the reliable 2D features. Conversely, the MI value increases at the PoseMoE Decoder stage (Layer 13). This high MI is proof that the CEKA mechanism has successfully integrated the purified, high-quality complementary knowledge from both experts. In summary, the MI analysis validates our design philosophy: using low-MI structural decoupling via the Encoder to solve contamination, followed by high-MI strategic aggregation via the Decoder to maximize knowledge complementarity. In summary, the MI curve perfectly validates the PoseMoE design philosophy: achieving low-MI structural decoupling via the MoE Encoder to solve the contamination problem, and then achieving high-MI strategic aggregation via the PoseMoE Decoder to maximize the value of knowledge complementarity.

\begin{figure}[t]
\centering
\includegraphics[width=1\linewidth]{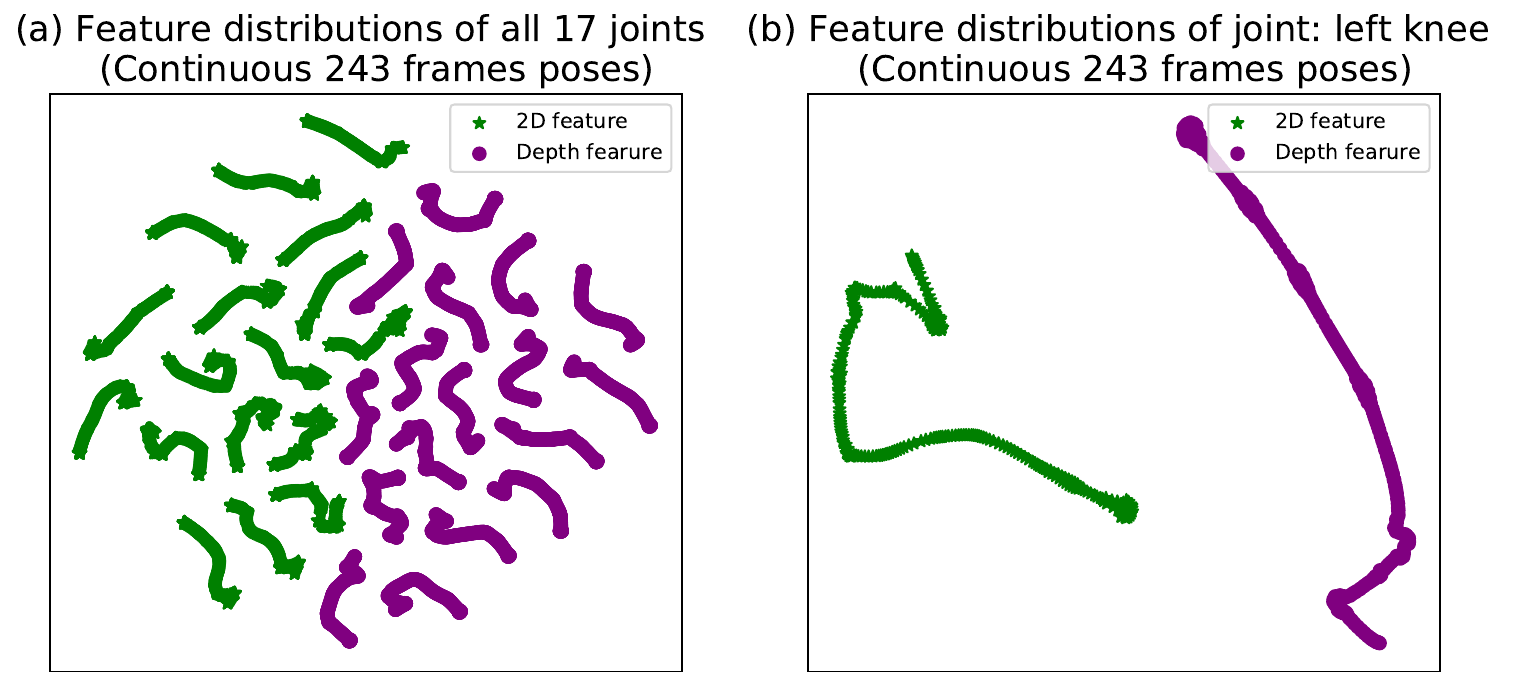}
\caption{Feature distributions visualization of 2D features (green) and depth features (purple) using t-SNE~\cite{tsne} method on Human3.6M~\cite{h36m} dataset.  We select continuous 243 frames pose from the action sequence "Directions" of subject 1 in Human3.6M~\cite{h36m} and visualize the features before regression heads. The distributions of the 2D features and depth features are different across various situations.
}
\label{fig:8}
\end{figure}

\begin{figure}[t]
\centering
\includegraphics[width=1\linewidth]{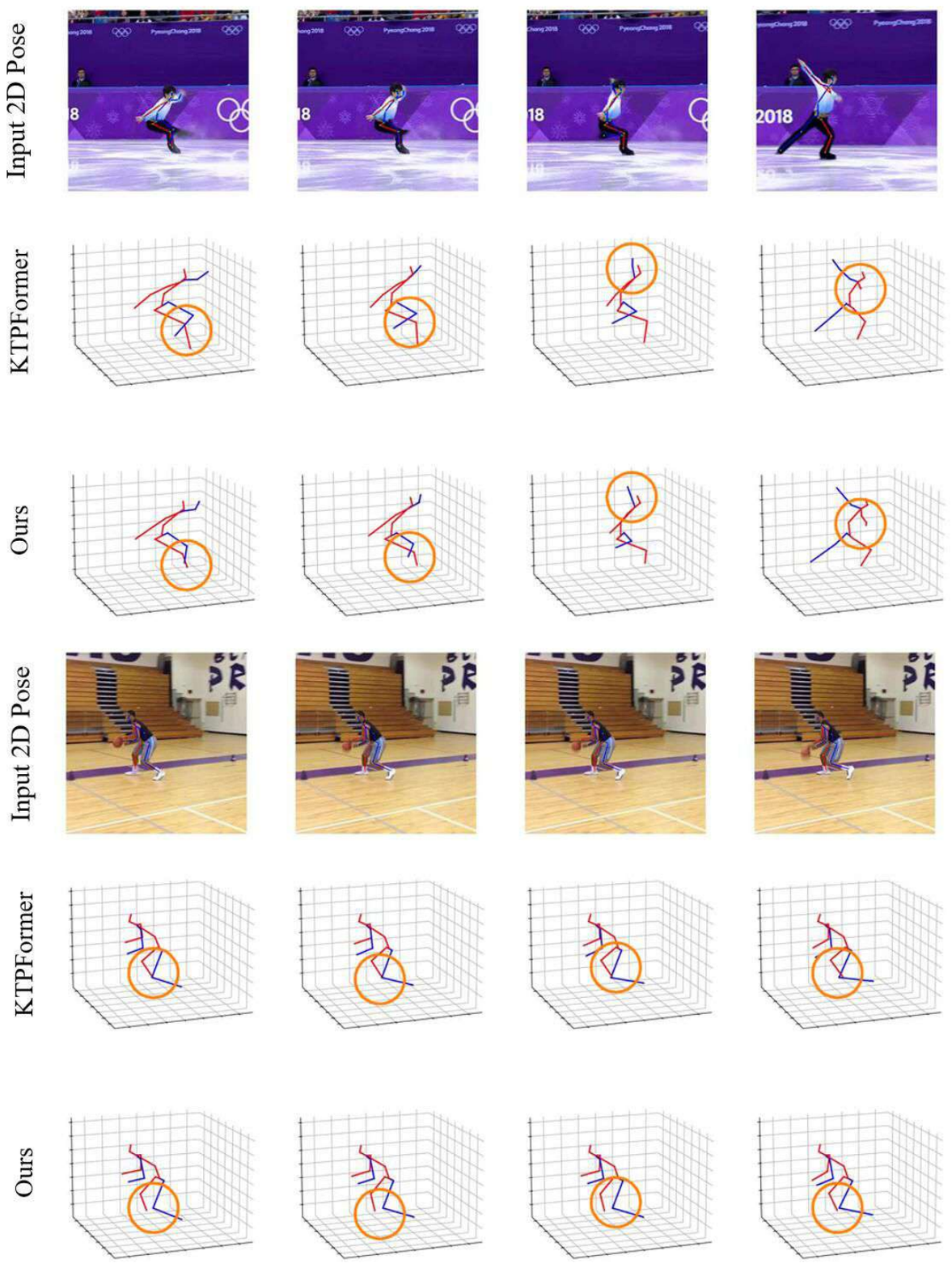}
\caption{Qualitative comparisons of PoseMoE with KTPFormer~\cite{peng2024ktpformer}. The green cycle indicates locations where our method achieves better results. 
haode% See Appendix~\ref{appendix:vis} for more comparison.
}
\label{fig:9}
\end{figure}

\subsection{Visualization}

\noindent\textbf{Cross Attention Visualization.}
As explicitly shown in Figure~\ref{fig:11} and Figure~\ref{fig:12}, the PoseMoE Decoder is emphatically not a simple feature fusion layer; instead, it meticulously complements global spatial-temporal dependencies via a bidirectional cross-attention mechanism operating across the distinct 2D pose feature and depth feature spaces, guaranteeing that only the most complementary and stabilized information is exchanged to maximize the final 3D pose estimation accuracy. Specifically, the fusion mechanism of the PoseMoE Decoder lies in utilizing the Bidirectional Cross-Attention mechanism to perform dynamic and complementary spatial and temporal knowledge transfer between the 2D pose features and depth features, thereby learning more comprehensive spatial joint dependencies and temporal motion relationships. Following the critical step where the PoseMoE Encoder decouples and initially refines the high-confidence 2D pose features and the low-confidence depth features into specialized representations, the PoseMoE Decoder is strategically deployed with the singular aim of aggregating complementary spatial-temporal knowledge between the two purified feature streams. The PoseMoE Decoder is demonstrably effective because it enables bidirectional and dynamic knowledge transfer, a process fundamentally more sophisticated than standard feature concatenation or simple fusion. Specifically, its architecture allows the two specialized experts—the 2D Pose Expert and the Depth Expert—to conditionally complement each other with information only after the necessary decoupling and refinement have taken place in the encoder. More precisely, the mechanism enables depth information to rigorously leverage the accurate spatial dependency and temporal consistency encoded within the 2D poses, which is vital for reducing its inherent scale ambiguity and stabilizing its predictions across sequences; simultaneously, it also allows the robust 2D poses to utilize the refined depth context information to strategically correct their projection errors in challenging scenarios involving partial occlusion or extreme viewing angles, where the initial 2D detection might be compromised.

\begin{figure}[t] 
\centering
\includegraphics[width=1\linewidth]{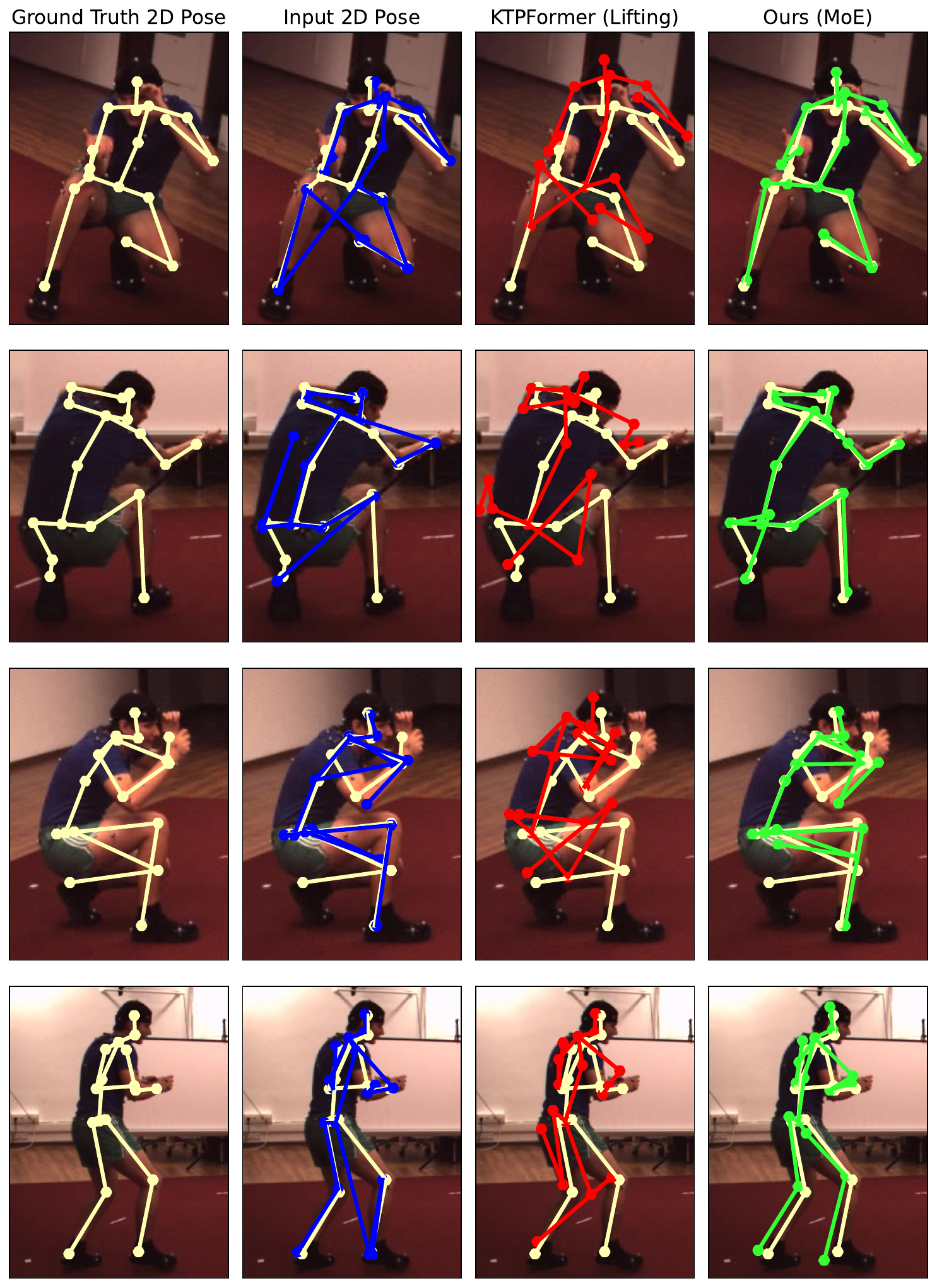}
% \vspace{-0.2cm}
\caption{Qualitative Comparison of 2D Pose (Ground Truth, Input, KTPFormer~\cite{peng2024ktpformer} and Ours). Our proposed PoseMoE produces more precise results in various challenging scenarios.
}
\label{fig:10}
\end{figure}

\begin{figure}[t] 
\centering
\includegraphics[width=1\linewidth]{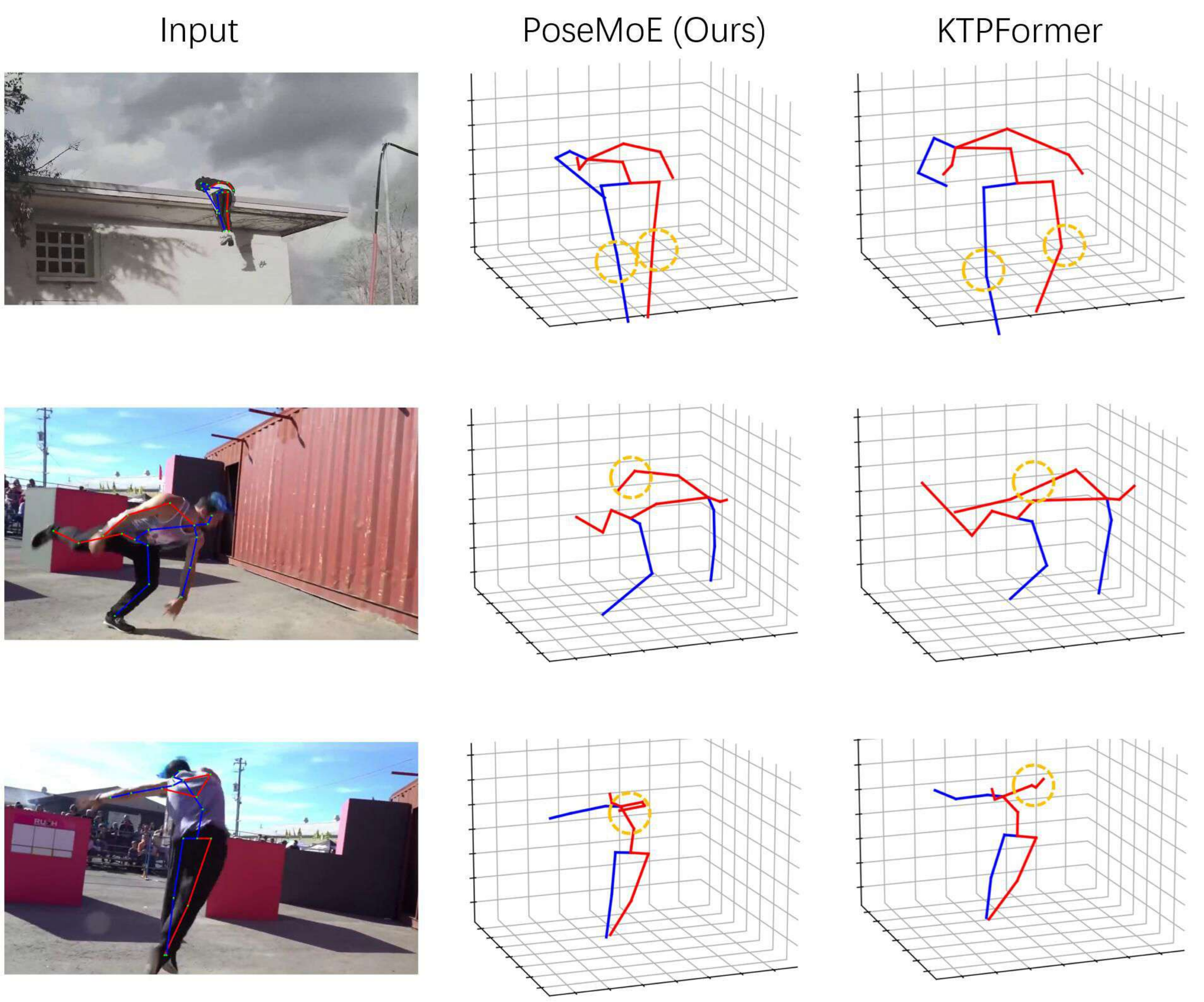}
\caption{Failure cases due to occlusion or motion blur.
}
\label{fig:14}
\end{figure}

\noindent\textbf{Feature Distribution Visualization.} To further analyze the difference between the 2D pose features and depth features, we utilize t-SNE~\cite{tsne} to visualize their feature distributions.  We select continuous 243 frames pose from action sequence "Directions" of subject 1 in Human3.6M~\cite{h36m} and visualize the features before regression heads (i.e., $\widetilde{F}_{2D}$ and $\widetilde{F}_{D}$). Figure~\ref{fig:8} exhibits distinct distributions between 2D features and depth features, which objectively demonstrates that these two types of features play different roles in the 3D human pose estimation process. This clear feature distribution separation, being a natural outcome of specialized learning, not only validates the critical necessity of rigorously disentangling 2D features and depth features during the essential feature learning process to prevent detrimental cross-contamination, but it also conclusively confirms that our novel proposed PoseMoE architecture, through its specialized Mixture-of-Experts design, is highly effective and capable of learning targeted, high-quality feature representations that are precisely optimized for the specific challenges and reliability characteristics of each respective domain.
%
% Due to space limitations, we present more visualization in our Appendix~\ref{appendix:vis}.

\noindent\textbf{3D Human Pose Estimation Visualization.}
We present 3D human pose estimation results by our proposed PoseMoE and KTFormer~\cite{peng2024ktpformer}. As shown in Figure~\ref{fig:9}, our method generalizes well to in-the-wild videos including self-occlusion and fast motion.

\noindent\textbf{2D Human Pose Estimation Visualization.}
We project the 2D pose in the camera coordinate system back to the image coordinate system for comparison. As shown in Figure~\ref{fig:10}, the lifting-based method~\cite{peng2024ktpformer} gets a 2D pose worse than the input, which contradicts our intuition. In contrast, our proposed PoseMoE obtains a 2D pose better than the input.

\noindent\textbf{Failure Cases.} There are some typical failure cases attributed to extreme occlusion or motion blur, where the 2D input may be unreliable, as shown in Figure~\ref{fig:14}. Although KTPFormer~\cite{peng2024ktpformer} and our PoseMoE all fail to produce fairly good estimations in these cases, our PoseMoE is capable of producing relatively more precise results.
\section{Conclusion}
\label{CONCLUSION}
This work presents a novel mixture-of-experts network named PoseMoE for monocular 3D human pose estimation. 
Our method addresses the limitation of the lifting-based framework that neglects different initial states between a well-detected 2D pose and an unknown depth. 
PoseMoE first learns 2D pose features and depth features separately by a mixture-of-experts architecture and then employs a PoseMoE decoder to indirectly supplement information between the refined 2D pose features and the learned
depth features. 
%
%
%Quantitative results provide empirical evidence that our PoseMoE can mitigate the erosion of the 2D pose caused by the uncertainty of depth features. 
%
Extensive quantitative experimental results on Human3.6M, MPI-INF-3DHP, and 3DPW datasets show that our PoseMoE outperforms the conventional lifting-based framework in terms of accuracy and robustness with fewer parameters. \\
%We hope our effort can provide a new framework for monocular 3D human pose estimation.
% \textbf{Limitation and Future Work.}
% \textcolor{blue}{\textbf{Limitation.} The core contribution of our work is providing a new framework for monocular 3D human pose estimation. To this end, we use the widely used spatial and temporal transformer as our encoder to ensure a fair comparison with the lifting-based framework. It will be novel and interesting to design specific encoders for different tasks to extend our framework in future research.}
\noindent\textbf{Future Work.} Although we propose a novel Mixture-of-Experts Network for monocular 3D human pose estimation, our input remains the same as conventional lifting-based methods: a 2D pose sequence. In scenarios where extreme occlusion or severe blurriness leads to unreliable outputs from the 2D pose detector, our disentanglement mechanism provides more accurate results, but it still cannot fully offset the systematic impact of highly erroneous 2D coordinates on 3D pose estimation. Beyond single-person settings, extending PoseMoE to multi-person, crowded, or extreme in-the-wild scenarios is a promising direction, allowing our disentangled MoE architecture to more robustly handle issues like severe occlusion and inter-person depth ambiguity by maintaining the integrity of the specialized 2D and depth features. In light of this, our future research will focus on integrating uncertainty estimation into the PoseMoE framework to enable more refined knowledge aggregation and extending PoseMoE to the multi-person setting.
% \textbf{Future Work.} The core contribution of our work is providing a new framework for monocular 3D human pose estimation. To this end, we use the widely used spatial and temporal transformer as our encoder to ensure a fair comparison with the lifting-based framework. It will be novel and interesting to design specific encoders for different tasks to extend our framework in future research. 

\bibliographystyle{IEEEtran}
\bibliography{refs}
% author
\begin{IEEEbiography}[{\includegraphics[width=1in,height=1.25in,clip,keepaspectratio]{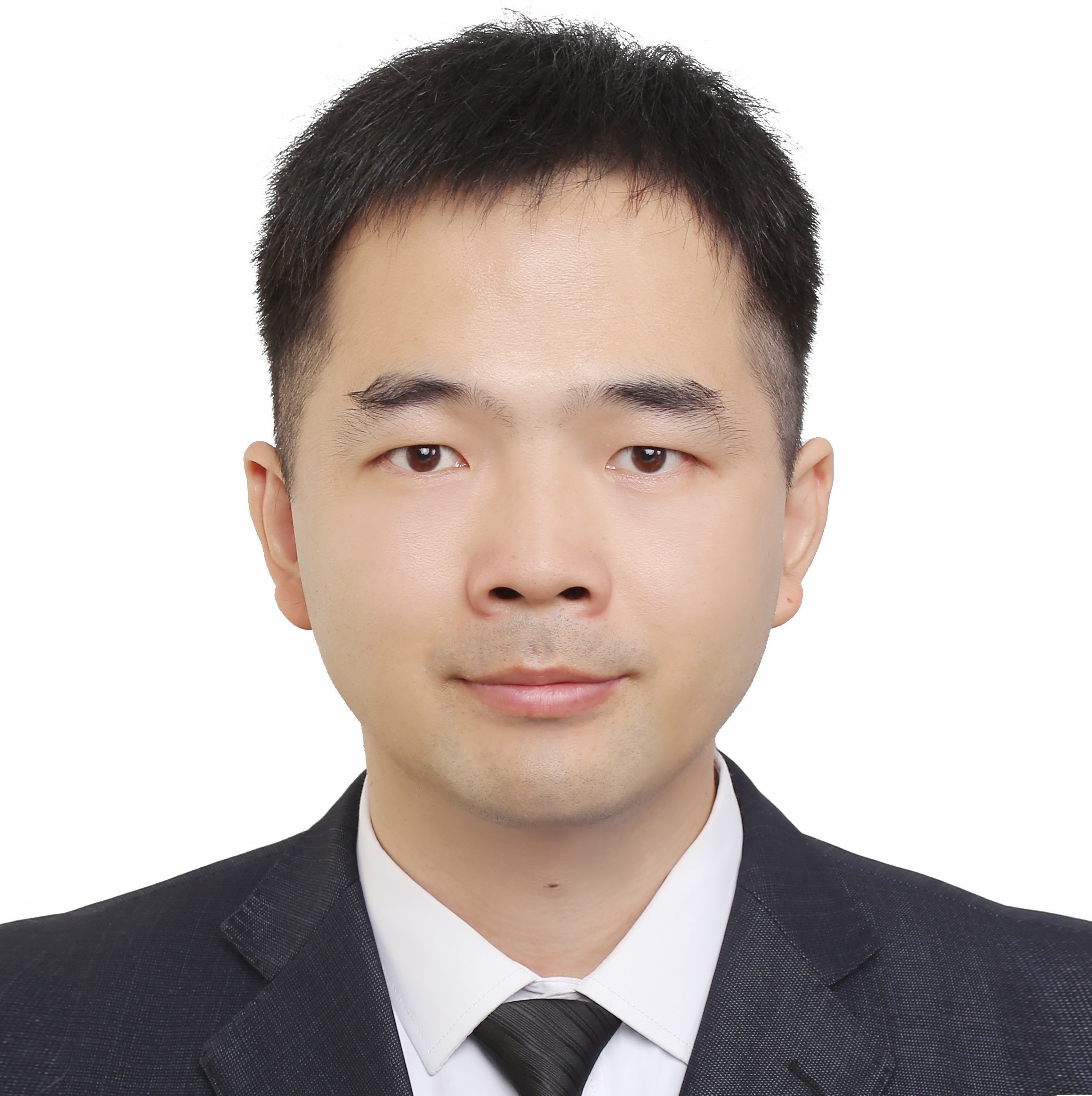}}]{Mengyuan Liu}
received his Ph.D. degree from the School of Electrical Engineering and Computer Science, Peking University. He was a research fellow at the School of Electrical and Electronic Engineering, Nanyang Technological University. Currently, he is an Assistant Professor at Peking University, Shenzhen Graduate School. His research focuses on human-centric perception and robot learning. His work has been published in leading conferences and journals, including CVPR and T-PAMI. He serves as an Associate Editor for PR and TIP.
\end{IEEEbiography}
\begin{IEEEbiography}[{\includegraphics[width=1in,height=1.25in,clip,keepaspectratio]{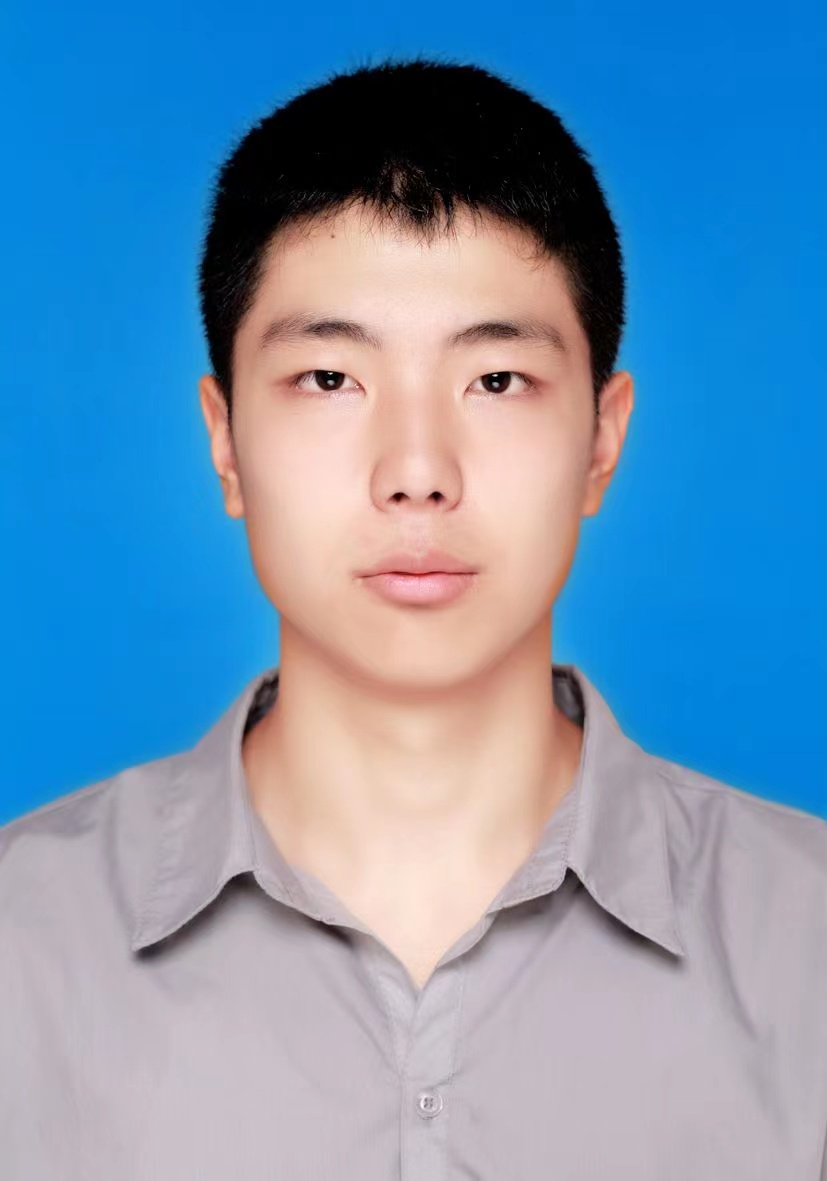}}]{Jiajie Liu} is currently pursuing the master’s degree with the School of Electronic and Computer Engineering, Peking University (PKU), China. His research interests include 3D human pose estimation and computer vision.
\end{IEEEbiography}
\begin{IEEEbiography}[{\includegraphics[width=1in,height=1.25in,clip,keepaspectratio]{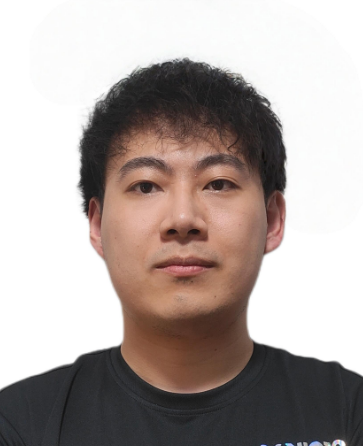}}]{Jinyan Zhang} received the B.S. degree from China University of Geosciences (CUG), China. He is a research graduate student studying at Peking University (PKU), China. His current research interest lies in 3D human pose and shape estimation.
\end{IEEEbiography}
\begin{IEEEbiography}[{\includegraphics[width=1in,height=1.25in,clip,keepaspectratio]{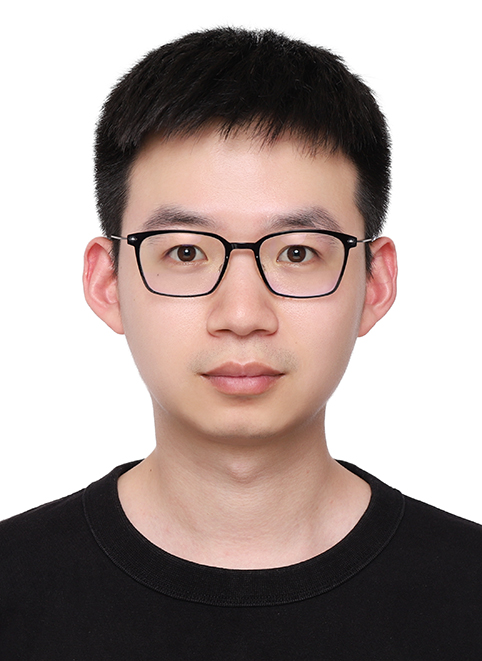}}]{Wenhao Li} is currently a Postdoctoral Researcher at the School of Computer Science and Engineering, Nanyang Technological University, Singapore.  
He received the Ph. D. degree from the School of Computer Science, Peking University, China. 
His research interests lie in deep learning, machine learning, and their applications to computer vision.
\end{IEEEbiography}
\begin{IEEEbiography}[{\includegraphics[width=1in,height=1.25in,clip,keepaspectratio]{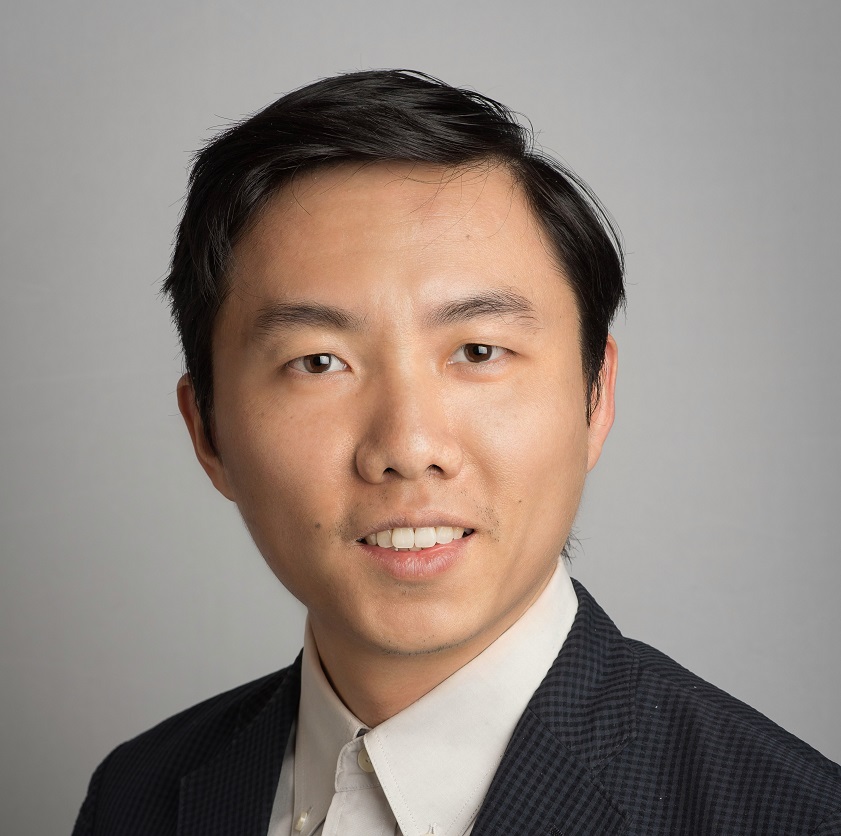}}]{Junsong Yuan} (Fellow, IEEE) received the B.Eng. degree from the Huazhong University of Science Technology (HUST) in 2002, the M.Eng. degree from the National University of Singapore in 2005, and the Ph.D. degree from Northwestern University in 2009. He is currently a Professor and the Director of the Visual Computing Laboratory, Department of Computer Science and Engineering (CSE), State University of New York at Buffalo, Buffalo, NY, USA. Before joining SUNY Buffalo, he was an Associate Professor (2015–2018) and Nanyang Assistant Professor (2009–2015) with Nanyang Technological University (NTU), Singapore. He is a fellow of IAPR.  He received the Chancellor’s Award for Excellence in Scholarship and Creative Activities from SUNY, Nanyang Assistant Professorship from NTU, the Outstanding EECS Ph.D. Thesis Award from Northwestern University, and the Best Paper Award from IEEE TRANSACTIONS ON MULTIMEDIA. He also serves as the General/Program Co-Chair for ICME and an Area Chair for CVPR, ICCV, ECCV, and ACM MM. He serves as a Senior Area Editor for \textit{Journal of Visual Communication and Image Representation} (JVCI) and an Associate Editor for IEEE TRANSACTIONS ON PATTERN ANALYSIS AND MACHINE INTELLIGENCE, IEEE TRANSACTIONS ON IMAGE PROCESSING, IEEE TRANSACTIONS ON CIRCUITS AND SYSTEMS FOR VIDEO TECHNOLOGY, and \textit{Machine Vision and Applications}. \end{IEEEbiography}

\end{document}